\documentclass[11pt]{article}

\usepackage[preprint]{acl}

\usepackage{times}
\usepackage{latexsym}

\usepackage[T1]{fontenc}

\usepackage[utf8]{inputenc}

\usepackage{microtype}

\usepackage{inconsolata}

\usepackage{graphicx}

\usepackage[T1]{fontenc}

\definecolor{snsblue}{RGB}{59, 117, 175}
\definecolor{snsgreen}{RGB}{81, 158, 62}
\definecolor{snsorange}{RGB}{239, 134, 53}
\definecolor{snsred}{RGB}{234, 51, 35}
\usepackage{xcolor}
\usepackage[table]{xcolor}

\newlength{\needcitelength}
\newcommand{\needcite}[1]{\settowidth{\needcitelength}{cites: {#1}}\todo[noinlinepar,inlinewidth=\the\needcitelength,size=\scriptsize]{cites: {#1}}}
\newcommand{\needref}[1]{\settowidth{\needcitelength}{ref: {#1}}\todo[color=yellow,noinlinepar,inlinewidth=\the\needcitelength,size=\scriptsize]{ref: {#1}}}

\usepackage{amsmath}
\usepackage{amssymb}
\usepackage{amsfonts}
\usepackage{bbm}

\usepackage[capitalise,english,nameinlink]{cleveref} %

\crefname{figure}{\text{Fig.}}{\text{Fig.}}
\crefname{section}{\S}{\S\S}
\crefname{equation}{\text{Eq.}}{\text{Eq.}}
\crefname{table}{\text{Tbl.}}{\text{Tbl.}}
\creflabelformat{equation}{#2\textup{#1}#3}
\crefname{appendix}{\mbox{App.}}{\mbox{App.}}

\usepackage{booktabs}
\usepackage{bm}
\usepackage{makecell}
\usepackage{multirow}

\usepackage{graphicx}

\title{Language Models Compare Quantities Using \\Number-specific and Unit-specific Heuristics}

\author{
Mutsumi Sasaki${}^{1}$ \quad
Go Kamoda${}^{2,3}$ \quad
Ryosuke Takahashi${}^{1,5,4}$ \quad
Kosuke Sato${}^{1}$ \quad
\\
{\bf Kentaro Inui${}^{5,1,4}$ \quad
Keisuke Sakaguchi${}^{1,4}$ \quad
Benjamin Heinzerling${}^{4,1}$ }
\\
${}^1$Tohoku University \quad
${}^2$SOKENDAI \quad
${}^3$NINJAL \quad
${}^4$RIKEN \quad
${}^5$MBZUAI\\
 \small{
   \textbf{Correspondence:} \href{mailto:mutsumi.sasaki@dc.tohoku.ac.jp}{mutsumi.sasaki@dc.tohoku.ac.jp}
 }
}

\begin{document}
\maketitle
\begin{abstract}
Quantities with measurement units, such as \textit{110 cm} and \textit{1.2 m}, require language models (LMs) to combine a numeral with a symbolic unit scale.
Here, we study how LMs compare such quantities in controlled settings spanning several unit systems.
We find that accuracy degrades near the comparison boundary, where small changes in value determine the correct answer.
The resulting errors are systematic: partial comparison cues derived separately from numerals and unit scales outperform alternative explanations in predicting LM preferences with linear surrogates and steering LM outputs through causal interventions.
The results suggest that LM behavior is better explained by a bag of heuristics account over numeral and unit-scale cues than by an account based solely on first converting both expressions into shared-scale representations.

\includegraphics[width=1em]{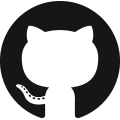}
\href{https://github.com/mutsumisasaki/LM-Interpret-Units}{mutsumisasaki/LM-Interpret-Units}

\end{abstract}

\section{Introduction}

Prior work has shown that language models (LMs) encode and compare numerical information in an interpretable manner.
Behavioral studies have found magnitude-comparison effects in LM representations~\citep{shah-etal-2023-numeric}, while mechanistic work has analyzed how LMs compute greater-than judgments and compare numeric properties~\citep{hanna2023how,el-shangiti-etal-2025-geometry}.
Other work has studied how numerical values or numeric properties are represented internally~\citep{heinzerling-inui-2024-monotonic,levy-geva-2025-language}.
However, quantities in text often appear with measurement units, such as \textit{110 cm}, \textit{1.2 m}, or \textit{3 kg}.
Comparing such quantities requires the LM to combine the numeral with the unit scale: \textit{110 cm} has a larger numeral than \textit{1.2 m}, but denotes a smaller length.
\begin{figure*}[t!]
\includegraphics[width=\linewidth]{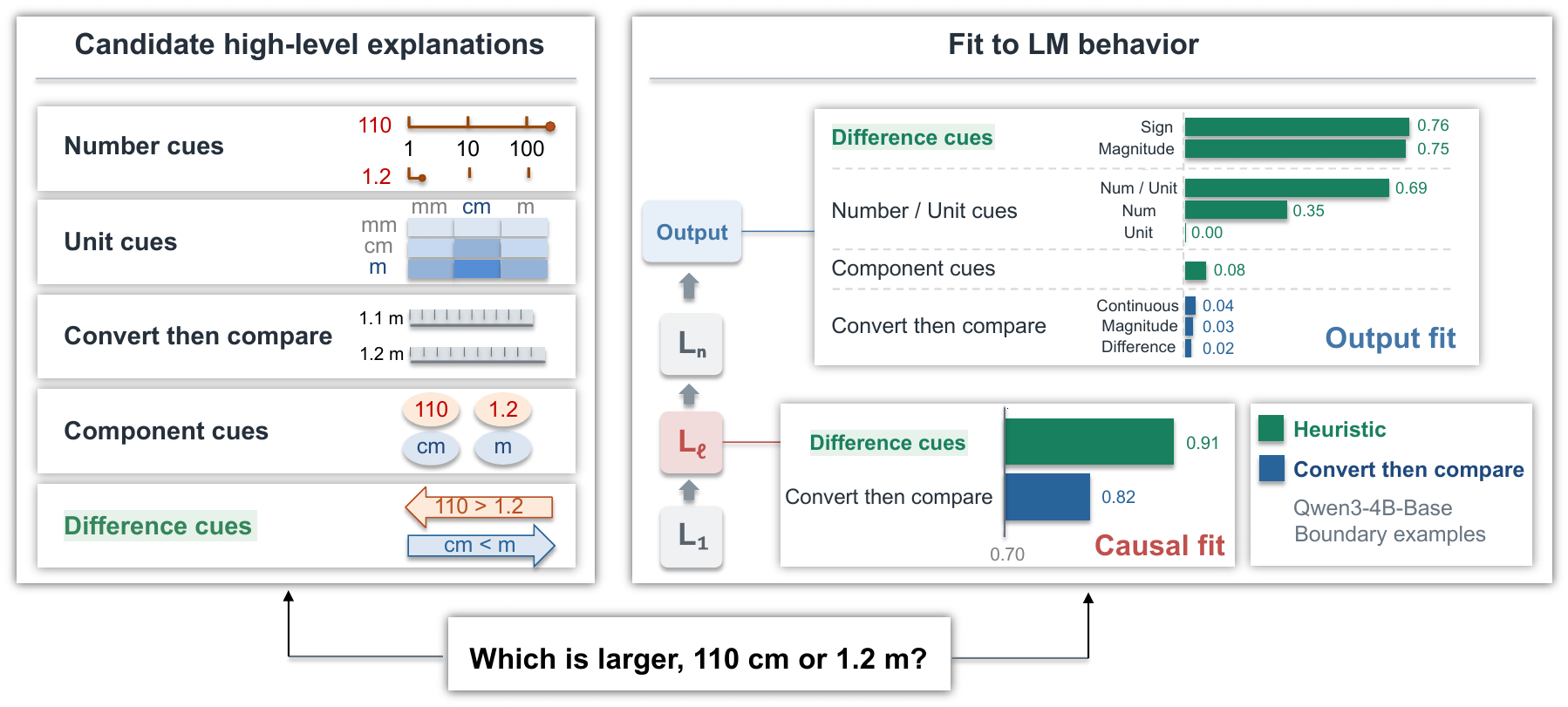}
\centering
\caption{
Overview of our approach and main findings.
For quantity comparison, we evaluate the relative fit of several candidate high-level explanations (left) to LM behavior at both the output and internal representation levels (right).
Our main finding is that LM behavior is better explained by aggregating numerical-difference and unit-scale-difference heuristics than by converting quantities to a common unit before comparing them.
}
\label{fig:figure1}
\end{figure*}
Here, we study comparison of quantities with measurement units as a setting where numerical comparison must be performed over heterogeneous expressions.
Unlike comparison of bare numbers, each expression contains both a numeral and a symbolic unit, and the larger numeral need not correspond to the larger quantity.
Quantity comparison also differs from prior interpretability work on fixed arithmetic functions over two numerals, including the four basic arithmetic operations such as addition, subtraction, multiplication, and division~\citep{stolfo-etal-2023-mechanistic, quirke-barez-2024-addition, quirke-etal-2025-addition-subtraction, zhang-etal-2024-arithmetic, zhou2024pretrained}, as well as modular addition~\citep{zhong2023the, nanda2023progress, ding2024survival}.
Here, the relevant operation is not a fixed arithmetic function over two numerals: the LM must combine each numeral with a symbolic unit and compare the resulting magnitudes on a shared scale.
The setting, therefore, gives a controlled test case for extending interpretability work from numerical operations to quantity representations in text.

Our results suggest that a heuristic account based on aggregating comparison cues derived separately from numerals and unit scales better captures LM behavior in quantity comparison than an account based on quantity representations that integrate numerals and unit scales on a shared scale.
Across multiple LMs and unit systems, accuracy decreases near the comparison boundary, where the two quantities are close in value.
To understand this behavior, we define several candidate explanations based on the structure of the quantity comparison task and evaluate their relative fit to LM behavior from a Causal Abstraction perspective~\citep{geiger2025ca}.
The candidate explanations we compare include accounts based on shared-scale representations of the two quantities, accounts based on separate representations and comparisons of their numeral and unit components, and accounts based on more surface-level heuristic cues such as digit count or specific unit pairs.
Among the candidates tested, a bag-of-heuristics account over numerical and unit-scale cues best captures LM behavior.

This parallels recent evidence that LMs solve arithmetic through a bag of heuristics~\citep{nikankin-etal-2025-arithmetic}.
By heuristics, we mean partial comparison cues that support a comparison decision.
Based on the structure of the quantity comparison task, we characterize a cue as heuristic if a comparison rule based on that cue alone does not agree with the correct comparison rule over the full input domain (see \cref{sec:definition_of_heuristic} for formal definitions).
\footnote{
Our definition of heuristics differs slightly from that of \citet{nikankin-etal-2025-arithmetic}: ours is based on task structure, whereas theirs is based on the activation patterns and functional roles of individual neurons (see \cref{sec:definition_of_heuristic} for details).
}

While the relevant heuristics for the four basic arithmetic operations include diverse and nontrivial task-solving cues, such as operand ranges, modulo patterns, and result ranges, numerical difference and unit-scale difference emerge as particularly important heuristic cues in our quantity-comparison setting.
These cues include whether the left numeral is larger, whether the left unit has a larger scale, and thresholded versions of these differences.
Aggregating heuristics then means that the LM preference can be predicted from a weighted combination of such cues.
This account can explain correct answers when the cues agree, but also predicts systematic errors when numeral and unit-scale cues point in different directions.

We test this account in three steps, summarized in \cref{fig:figure1}.
First, we measure behavioral accuracy across controlled quantity comparisons and show that performance degrades near the comparison boundary.
Second, we fit surrogate models corresponding to the candidate explanations and find that numerical-difference and unit-scale-difference features predict LM preferences better than the alternative candidates we test.
Third, we use causal interventions and find that interventions on subspaces aligned with numerical difference and unit-scale difference more effectively shift the LM output in the corresponding direction than interventions based on the alternative variables.
Together, these results suggest that heuristic aggregation over numerical and unit-scale cues provides an effective account for understanding LM behavior in quantity comparison.

\begin{table*}[t]
\centering
\small
\begin{tabular}{llp{0.48\linewidth}}
\toprule
\textbf{Unit setting} & \textbf{Physical dimension \(D\)} & \textbf{Unit set \(U_D\)} \\
\midrule
Metric length
&
\(\mathrm{length}\)
&
\(\{\mathrm{mm}, \mathrm{cm}, \mathrm{m}, \mathrm{km}\}\)
\\
Imperial length
&
\(\mathrm{length}\)
&
\(\{\mathrm{inch}, \mathrm{feet}, \mathrm{yard}, \mathrm{mile}\}\)
\\
Metric mass
&
\(\mathrm{mass}\)
&
\(\{\mathrm{mg}, \mathrm{g}, \mathrm{kg}, \mathrm{t}\}\)
\\
Metric-imperial length
&
\(\mathrm{length}\)
&
\(\{\mathrm{mm}, \mathrm{cm}, \mathrm{m}, \mathrm{km}, \mathrm{inch}, \mathrm{feet}, \mathrm{yard}, \mathrm{mile}\}\)
\\
Metric-imperial mass
&
\(\mathrm{mass}\)
&
\(\{\mathrm{g}, \mathrm{kg}, \mathrm{ounce}, \mathrm{pound}\}\)
\\
\bottomrule
\end{tabular}
\caption{
Unit settings used in our experiments.
Each setting is defined by a physical dimension \(D\) and a mutually comparable unit set \(U_D\).
}
\label{tab:unit_settings}
\end{table*}

\section{Task Formulation and Overall Experimental Setup}
\label{sec:task_formulation}
To analyze how LMs combine numerical values with symbolic unit information, we study the comparison of quantities with measurement units as a controlled task.
This section describes the formulation of the comparison task over quantities with measurement units, as well as the experimental setup shared across this study.

\subsection{Task Formulation}
\label{sec:task_fomulation}
Let \(D\) denote a physical dimension, such as length or mass, and let \(U_D\) be a chosen set of mutually comparable measurement units for that dimension.
A quantity with a measurement unit is specified by a numerical value \(r\in[r_{\min}, r_{\max}]\subset\mathbb{R}_{>0}\) and a unit \(u\in U_D\), and is written as \(ru\).
Given two quantities \(q_1=r_1u_1\) and \(q_2=r_2u_2\), where \(u_1,u_2\in U_D\), the LM is asked to decide which is larger:
\begin{equation}
    \text{Which is larger, } 
    r_1 u_1
    \text{ or }
    r_2 u_2?
\label{eq:input}
\end{equation}

For example, in the comparison ``Which is larger, 110 cm or 1.2 m?'', we have \(D=\mathrm{length}\), \(U_D=\{\mathrm{mm}, \mathrm{cm}, \mathrm{m}, \mathrm{km}\}\), \(r_1=110\), \(u_1=\mathrm{cm}\), \(r_2=1.2\), and \(u_2=\mathrm{m}\).

For each comparison \(q_1=r_1u_1\) and \(q_2=r_2u_2\), the correct answer can be determined by the log-scale difference between the two quantities after accounting for unit scale:
\begin{equation}
\begin{aligned}
    QM(q_1,q_2)
    &=
    \log(r_1)
    -
    \log(r_2) \\
    &+
    \log(s_D(u_1))
    -
    \log(s_D(u_2)).
\end{aligned}
\label{eq:quantity_margin}
\end{equation}
Here, \(s_D(u)\) denotes the scale of unit \(u\) relative to a base unit for dimension \(D\).
We refer to \(QM(q_1,q_2)\) as the Quantity Margin. Throughout this paper, we use the Quantity Margin to quantify how close a comparison is to the decision boundary and to analyze LM behavior systematically across different numerical values and unit pairs.
We formulate the Quantity Margin in the log space, following prior work~\citep{alquabeh2026number} and our preliminary experiments under the unit settings used in this study (\cref{sec:motivation_qm}).

\(q_1\) is larger when \(QM(q_1,q_2)>0\), and the right quantity \(q_2\) is larger when \(QM(q_1,q_2)<0\).

For the example above, taking meters as the base unit gives \(s_D(\mathrm{cm})=10^{-2}\) and \(s_D(\mathrm{m})=1\).
Thus,
$
QM(110\mathrm{cm},1.2\mathrm{m})
=
\log(110)
-
\log(1.2)+
\log(10^{-2})
-
\log(1)
\approx
-0.04 .
$

The negative margin indicates that \(1.2\mathrm{m}\) is larger, and its small absolute value indicates that the comparison is close to the decision boundary.

\subsection{Overall Experimental Setup}
\label{sec:overall_experimental_setup}

We consider five unit settings, each defined by a physical dimension \(D\) and a mutually comparable unit set \(U_D\), as shown in \cref{tab:unit_settings}.
For numerical values, we sample \(r\) from the range \(10^{-3} \le r < 10^4\) across all unit settings.
The first three settings contain units from a single unit system, while the last two combine metric and imperial units, allowing us to test whether the observed behavior is specific to a single unit system or generalizes across both homogeneous and heterogeneous unit comparisons.

We evaluate five base LMs (Qwen3-4B-Base, Qwen3-8B-Base, Qwen3-14B-Base~\citep{yang2025qwen3technicalreport}, Olmo-3-1025-7B, and Olmo-3-1125-32B~\citep{olmo2026olmo3}) and two instruct LMs (Qwen3-4B and Qwen3-8B). 
We use the default tokenizer provided with each LM, and represent all numerical values in standard decimal notation (i.e., without scientific notation).

We use multiple prompt templates that ask the LMs to output the larger quantity itself, use greedy decoding, and evaluate predictions by exact match with the correct quantity.
Further details on prompt templates, unit scales, tokenization, and evaluation are provided in \cref{sec:appendix_experimental_setup}.

\section{Behavioral Observation: LMs Become Less Accurate Near the Boundary}
\label{sec:behavioral_analysis}
In this section, we examine at the behavioral level under what conditions LMs succeed or fail at comparing quantities with measurement units.

\subsection{Experimental Setup}
\label{sec:behavioral-setup}
To analyze LM behavior systematically, we evaluate their accuracy on the quantity comparison task defined in \cref{sec:task_formulation}.
We create a controlled dataset by varying both the numerical values and the unit pairs.
We group 5,000 comparison examples into 18 bins according to their Quantity Margin, defined in \cref{sec:task_formulation}, resulting in 277 or 278 examples per bin.
The bins cover negative and positive margins symmetrically, with finer intervals near the decision boundary, e.g., \([-0.2,0)\) and \([0,0.2)\).
Using the LMs, prompt templates, and unit settings described in \cref{sec:task_formulation}, we then evaluate accuracy within each bin.

\subsection{Results}
\begin{figure}[t!]
\includegraphics[width=\linewidth]{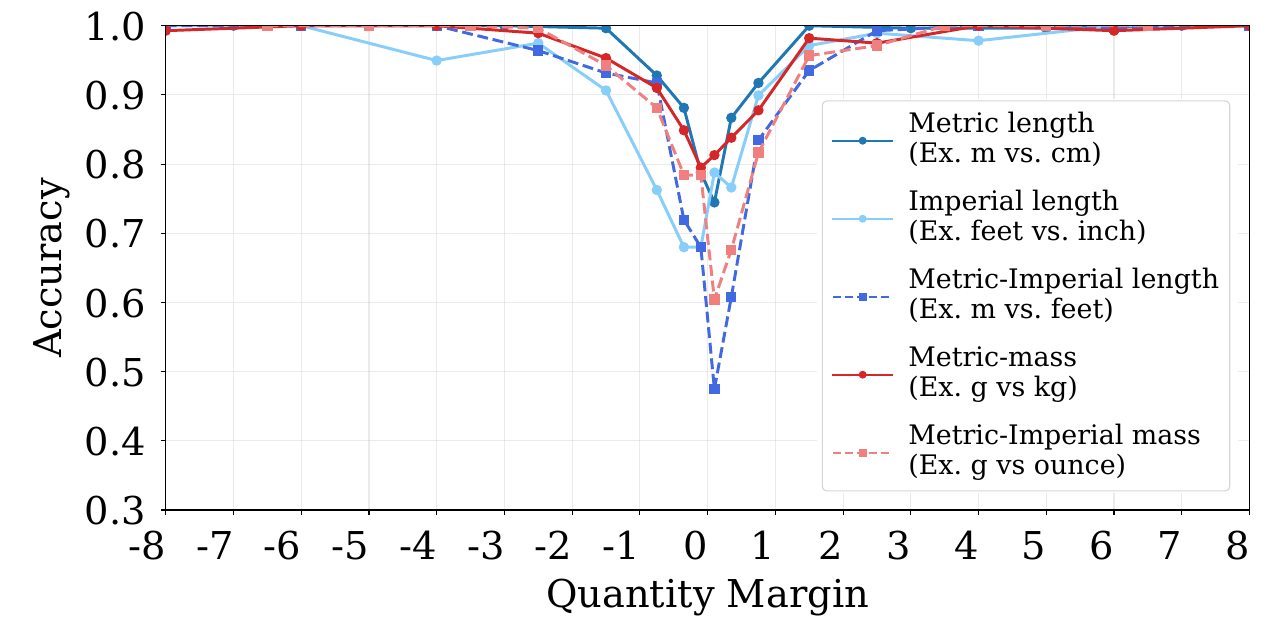}
\centering
\caption{
Accuracy of comparisons between quantities with measurement units, grouped by Quantity Margin (Qwen3-4B-Base).
Blue and red denote length and mass comparisons, respectively.
Dashed curves denote heterogeneous metric-imperial comparisons.
}
\label{fig:behavioral_analysis_numunit}
\end{figure}
\cref{fig:behavioral_analysis_numunit} shows the result for Qwen3-4B-Base across the five unit settings.
The LM's accuracy is strongly organized by the Quantity Margin.
Accuracy remains high when the absolute margin is large, but decreases sharply as examples approach the comparison boundary.
This indicates that comparisons are easy for LMs when the Quantity Margin is large and become gradually more difficult as the margin approaches zero.
The same margin-dependent pattern appears across length and mass comparisons, as well as across homogeneous and heterogeneous unit comparisons.
At the same time, heterogeneous metric-imperial comparisons tend to be less accurate than comparisons within a single unit system, with metric length comparisons generally easier than imperial length comparisons.
Additional results in \cref{sec:appendix_behavioral} show that the margin-dependent pattern is robust across LMs, prompt templates, and unit notations, while overall accuracy varies with prompt wording and unit surface form, such as \(m\) vs. \(meter\).

\section{Cue-Combination Hypothesis: Quantity Decisions Depend on Cue Agreement}
\label{sec: cue-agreement-hypothesis}
\begin{figure}[t!]
\includegraphics[width=\linewidth]{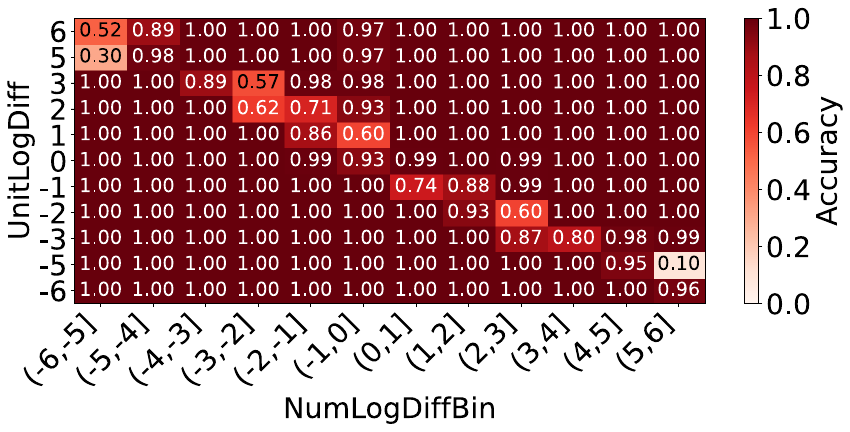}
\centering
\caption{
Accuracy of comparisons between quantities with measurement units, grouped by NumLogDiff and UnitLogDiff (Qwen3-4B-Base).
}
\label{fig:behavioral_heatmap_numunit_qwen3_4b}
\end{figure}
We interpret the drop in LM accuracy near the Quantity Margin boundary shown in \cref{fig:behavioral_analysis_numunit} through a cue-combination hypothesis, under which LM's preferences can be explained by a weighted combination of multiple comparison cues.
Under this view, comparison difficulty depends on how consistently the cues favor the same answer.
Far from the boundary, the cues tend to support the same answer and reinforce one another, whereas near the boundary, this support becomes weaker or less consistent.

The cue-combination view raises the question of which cues actually explain its behavior.
We therefore first enumerate candidate cues that could support quantity comparison, and then test their predictive power in the surrogate analysis in \cref{sec:surrogate_analysis}.

Several comparison cues are naturally available in this task.
The numerical-difference and unit-scale cues are defined as
\begin{equation}
\mathrm{NumLogDiff}=\log r_1-\log r_2,
\end{equation}
\begin{equation}
\mathrm{UnitLogDiff}=\log s_D(u_1)-\log s_D(u_2).
\end{equation}
The magnitude of each quantity on a common scale can also serve as a cue:
\begin{equation}
\mathrm{GlobalLogX}=\log r_1+\log s_D(u_1),
\end{equation}
\begin{equation}
\mathrm{GlobalLogY}=\log r_2+\log s_D(u_2).
\end{equation}

\begin{table*}[t]
    \centering
    \small
    \rowcolors{2}{gray!10}{white}
    \begin{tabular}{m{0.23\linewidth}m{0.2\linewidth}m{0.4\linewidth}}
        \toprule
        \textbf{Surrogate family} & \textbf{Cue variables} & \textbf{Role in analysis} \\
        \midrule
        
        Quantity Margin baseline
        &
        Quantity Margin
        &
        Tests how well the variable that determines the correct outcome predicts LM behavior on its own.
        \\[.2em]
        
        Global-quantity cues
        &
        \makecell[l]{GlobalLogX,\\ GlobalLogY}
        &
        Tests whether LM behavior is explained by the two quantities represented on a common scale.
        \\
        
        Primitive input components
        &
        \makecell[l]{\(\log r_1,\ \log r_2\),\\ \(\log s_D(u_1),\ \log s_D(u_2)\)}
        &
        Tests whether LM behavior is explained by the original numerical and unit components before forming comparison variables.
        \\
        
        Number/unit difference cues
        &
        \makecell[l]{NumLogDiff,\\ UnitLogDiff}
        &
        Tests whether LM behavior is explained by numerical and unit-scale differences.
        \\

        Number/unit pair cues
        &
        \makecell[l]{NumLogDiff,\\ \(\log s_D(u_1),\ \log s_D(u_2)\)}
        &
        Tests whether LM behavior is explained by numerical differences and unit-pair cues.
        \\
        
        All heuristic cues
        &        \makecell[l]{All heuristic cue variables\\ provided in \cref{tab:heuristic_features}}
        &
        Baseline for how well a broad set of heuristic cues can explain LM behavior without directly using the Quantity Margin.
        \\[.2em] 
        \bottomrule
    \end{tabular}
    \caption{
    Representative surrogate families used in the main analysis.
    Each surrogate predicts the LM log-probability margin from one family or combination of cue families.
    The full list of individual features is provided in \cref{tab:heuristic_features}, and the full list of surrogate models is provided in \cref{tab:surrogate_feature_sets}.
    }
    \label{tab:surrogate_representative_models}
\end{table*}

In addition to these comparison-related quantities, LMs may also rely on surface-level heuristic cues derived from the input representation, including numerical appearance,
\begin{equation}
\mathrm{firstdigit}(r_1)>n,
\end{equation}
where \(n\) is a threshold, unit identities,
\begin{equation}
u_2 \in \{\mathrm{km}, \mathrm{kg}, \mathrm{mile}, \ldots\},
\end{equation}
and familiar unit-pair patterns,
\begin{equation}
(u_1,u_2)=(\mathrm{cm},\mathrm{m}) \Rightarrow q_2.
\end{equation}

As a concrete example of cue agreement and conflict, we group examples by \(\mathrm{NumLogDiff}\) and \(\mathrm{UnitLogDiff}\).
For metric length comparisons, we bin \(\mathrm{NumLogDiff}\) over \([-6,6]\) with width 1, and use the discrete \(\mathrm{UnitLogDiff}\) values determined by unit pairs, \(\{-6,-5,-3,-2,-1,0,1,2,3,5,6\}\).
We generate 200 examples for each grid cell and evaluate LM accuracy in each cell.
\cref{fig:behavioral_heatmap_numunit_qwen3_4b} shows that errors concentrate near the diagonal region where numerical-difference and unit-scale cues cancel each other out, yielding a small Quantity Margin.
This pattern illustrates the cue-combination view: comparisons are easy when the relevant cues jointly support the same side, and difficult when they are weak or conflicting.
The same qualitative pattern appears across the other unit settings, as shown in \cref{sec:appendix_hypothesis}.

\section{Surrogate Analysis: Number and Unit Differences Best Predict LM Behavior}
\label{sec:surrogate_analysis}
In this section, we test which cues best predict the LMs' actual behavior.

\begin{table*}[t]
\centering
\small
\begin{tabular}{l|rrr|rrr}
\toprule
\multirow{2}{*}{\textbf{Feature set}} &
\multicolumn{3}{c|}{\textbf{All examples}} &
\multicolumn{3}{c}{\textbf{Boundary examples}} \\
&
\(R^2_{\mathrm{S}}\) & \(\Delta R^2\) & \(R^2_{\mathrm{partial}}\) &
\(R^2_{\mathrm{S}}\) & \(\Delta R^2\) & \(R^2_{\mathrm{partial}}\) \\
\midrule
All heuristic features & 0.886 & 0.196 & 0.634 & 0.687 & 0.666 & 0.680 \\
NumLogDiff \& UnitLogDiff thresholds & 0.836 & 0.147 & 0.472 & 0.749 & 0.728 & 0.740 \\
Signed NumLogDiff \& UnitLogDiff & 0.817 & 0.128 & 0.439 & 0.763 & 0.741 & 0.695 \\
NumLogDiff thresholds + unit-pair thresholds & 0.815 & 0.125 & 0.403 & 0.693 & 0.672 & 0.683 \\
Primitive component thresholds & 0.714 & 0.025 & 0.091 & 0.077 & 0.055 & 0.072 \\
GlobalLogX \& GlobalLogY thresholds & 0.713 & 0.023 & 0.087 & 0.027 & 0.006 & 0.024 \\
NumLogDiff \& UnitLogDiff & 0.712 & 0.022 & 0.071 & 0.063 & 0.042 & 0.043 \\
GlobalLogX \& GlobalLogY & 0.692 & 0.002 & 0.007 & 0.035 & 0.013 & 0.013 \\
Quantity Margin & 0.690 & 0.000 & 0.000 & 0.022 & 0.000 & 0.000 \\
\bottomrule
\end{tabular}
\caption{
Representative surrogate results for the metric length setting in Qwen3-4B-Base. 
The trained surrogate models are evaluated on All examples and Boundary examples. 
\(R^2_{\mathrm{QM}}\), \(R^2_{\mathrm{S}}\), and \(R^2_{\mathrm{QM+S}}\) denote the test \(R^2\) of the Quantity Margin baseline, the current surrogate model, and the surrogate model obtained by adding the Quantity Margin to the current feature set, respectively.
\(\Delta R^2=R^2_{\mathrm{S}}-R^2_{\mathrm{QM}}\), and \(R^2_{\mathrm{partial}}=(R^2_{\mathrm{QM+S}}-R^2_{\mathrm{QM}})/(1-R^2_{\mathrm{QM}})\). 
Rows are sorted by \(R^2_{\mathrm{partial}}\) on All examples.
}
\label{tab:meter_AbsGlobalDiffBin_hard_abs_lt_02_surrogate_representative_results}
\end{table*}

\subsection{Experimental Setup}
\label{sec:surrogate_setup}
We use surrogate models to test which candidate quantities and cues explain the LM's comparison behavior.
For each comparison example, we instantiate a prompt \(\mathrm{Prompt}(q_1,q_2)\), such as
``Which is larger, \(q_1\) or \(q_2\)? The answer is''.
We then compare the LM's average log-probability of generating the two candidate answer strings, namely the left quantity \(a_{q_1}\) and the right quantity \(a_{q_2}\), as continuations of this prompt:
\begin{equation}
\begin{aligned}
    m_{\mathrm{LM}}(q_1,q_2)
    &=
    \bar{\ell}_{\theta}(a_{q_1} \mid \mathrm{Prompt}(q_1,q_2))\\
    &-
    \bar{\ell}_{\theta}(a_{q_2} \mid \mathrm{Prompt}(q_1,q_2)).
\end{aligned}
\end{equation}
Here, \(\bar{\ell}_{\theta}(a \mid \mathrm{Prompt})\) denotes the average token log-probability of the full candidate string \(a\) when appended to the prompt.
A positive value indicates that the LM prefers the left quantity, while a negative value indicates that it prefers the right quantity.

\cref{tab:surrogate_representative_models} summarizes the representative surrogate families used in the main analysis.
These families are based on a systematic feature design described in \cref{sec:appendix_surrogate_features}.
Starting from the four primitive variables that determine the Quantity Margin, \(r_1\), \(s_D(u_1)\), \(r_2\), and \(s_D(u_2)\), we construct feature families corresponding to primitive components, number/unit differences, shared-scale global quantities.
We further instantiate these cues as signed, thresholded, and continuous features to capture heuristics at different levels of granularity.
The representative families are chosen to compare the main candidate explanations of LM behavior: the Quantity Margin, global quantities, primitive numerical and unit components, number/unit differences, and broad heuristic cue sets.
Each surrogate predicts the LM log-probability margin from one cue family or a combination of cue families.
The full list of individual features is provided in \cref{tab:heuristic_features}, and the full list of surrogate models is provided in \cref{tab:surrogate_feature_sets}.

We evaluate each surrogate model using its held-out coefficient of determination, denoted by \(R^2_{\mathrm{S}}\).
As a reference, we also evaluate a baseline surrogate that uses only the Quantity Margin, the variable that determines the correct comparison outcome, and denote its held-out coefficient of determination by \(R^2_{\mathrm{QM}}\).
Furthermore, for each surrogate feature set, we evaluate a surrogate model obtained by adding the Quantity Margin to that feature set, yielding \(R^2_{\mathrm{QM+S}}\).

Using these scores, we evaluate each surrogate model with two metrics, \(\Delta R^2\) and \(R^2_{\mathrm{partial}}\).
The metric $\Delta R^2$ measures the improvement in held-out predictive performance over the Quantity Margin baseline.
The partial $R^2$ quantifies the proportion of variance left unexplained by the Quantity Margin baseline that is additionally explained when the candidate feature set is added.
The metric \(R^2_{\mathrm{partial}}\) measures the proportion of the variance left unexplained by the Quantity Margin baseline that is additionally explained by the surrogate feature set.

\begin{equation}
    \Delta R^2
    =
    R^2_{\mathrm{S}}
    -
    R^2_{\mathrm{QM}}
\end{equation}

\begin{equation}
    R^2_{\mathrm{partial}}
    =
    \frac{R^2_{\mathrm{QM+S}}-R^2_{\mathrm{QM}}}
         {1-R^2_{\mathrm{QM}}}
\end{equation}

We further evaluate the surrogate models on the subset of boundary examples, where \(|QM(q_1,q_2)| \leq 0.2\).
As shown in \cref{fig:behavioral_analysis_numunit}, this is the region where LM accuracy drops most clearly.
Under the cue-combination hypothesis, this region is also where different cues are expected to become less consistently aligned.
Analyzing these boundary examples therefore allows us to test which feature sets best explain LM behavior when different cues are in conflict.

\paragraph{Implementation details.}
We construct a broad surrogate-analysis dataset by sampling examples evenly across quantity-margin bins.
We use 20,000 examples for training, 1,000 examples for validation, and 5,000 examples for testing.
The LM outputs used as regression targets are obtained with the same LMs, unit settings, prompt templates, and decoding setup as in \cref{sec:overall_experimental_setup}.
For each feature set, we fit a ridge regression surrogate model to predict \(m_{\mathrm{LM}}(q_1,q_2)\).
All features are standardized using the training set statistics.
The regularization strength is selected separately for each surrogate model from
\(\alpha \in \{0.01, 0.1, 1, 10, 100\}\)
based on validation \(R^2\), and the selected surrogate model is evaluated on the held-out test set.

\subsection{Results}
\label{sec:surrogate-results}
\cref{tab:meter_AbsGlobalDiffBin_hard_abs_lt_02_surrogate_representative_results} shows results for the representative surrogate families summarized in \cref{tab:surrogate_representative_models}, using Qwen3-4B-Base on metric length comparisons.
Here, signed features use only the direction of a cue, such as whether \(\mathrm{NumLogDiff}>0\) or \(\mathrm{UnitLogDiff}>0\), while threshold features use coarse magnitude information, such as whether a log-scale variable exceeds a threshold \(n\).
Other continuous feature sets use the original log-scale values directly.

For the all examples setting, the LM's behavior is well explained by heuristic features based on \(\mathrm{NumLogDiff}\) and \(\mathrm{UnitLogDiff}\).
The thresholded and signed NumLogDiff/UnitLogDiff feature sets achieve high $\Delta R^2$ and partial $R^2$ values among the representative models.
This indicates that the LM's log-probability margin is especially well captured by coarse cues about which side has the larger numeral and which side has the larger unit scale.
By contrast, feature sets based on $\mathrm{GlobalLogX}$, $\mathrm{GlobalLogY}$, or the primitive input components provide only limited improvements over the Quantity Margin baseline.
These results suggest that LM behavior is better explained by heuristic cues derived from numerical difference and unit-scale difference than by these alternative feature representations.

We next focus on boundary examples, where \(|QM(q_1,q_2)| \leq 0.2\), the region where LM accuracy drops most clearly in \cref{fig:behavioral_analysis_numunit}.
As expected, the Quantity Margin alone provides little explanatory power for the LM's behavior on these boundary examples.
Indeed, the Quantity Margin baseline achieves only \(R^2_{\mathrm{QM}} = 0.022\) in this region.
In contrast, our \(\mathrm{NumLogDiff}\) and \(\mathrm{UnitLogDiff}\) feature set achieves even higher \(\Delta R^2\) and \(R^2_{\mathrm{partial}}\) than on the full evaluation set.
This suggests that, in cue-conflicting cases where the LM frequently fails, its behavior remains well predicted by heuristic features based on numerical difference and unit-scale difference.

Full results across LMs, unit settings are provided in \cref{sec:appendix_surrogate_analysis}, where the same overall trend is consistently observed.
The results for all surrogate models are reported in \cref{tab:meter_surrogate_detailed_results}.

\section{Causal Analysis: Number and Unit Differences Steer LM Decisions}
\label{sec:internal_mechanism}
In this section, we test whether causal interventions on \(\mathrm{NumLogDiff}\) and \(\mathrm{UnitLogDiff}\) explain and steer LM behavior more effectively than interventions based on alternative candidate variables.

\begin{figure}[t!]
\includegraphics[width=\linewidth]{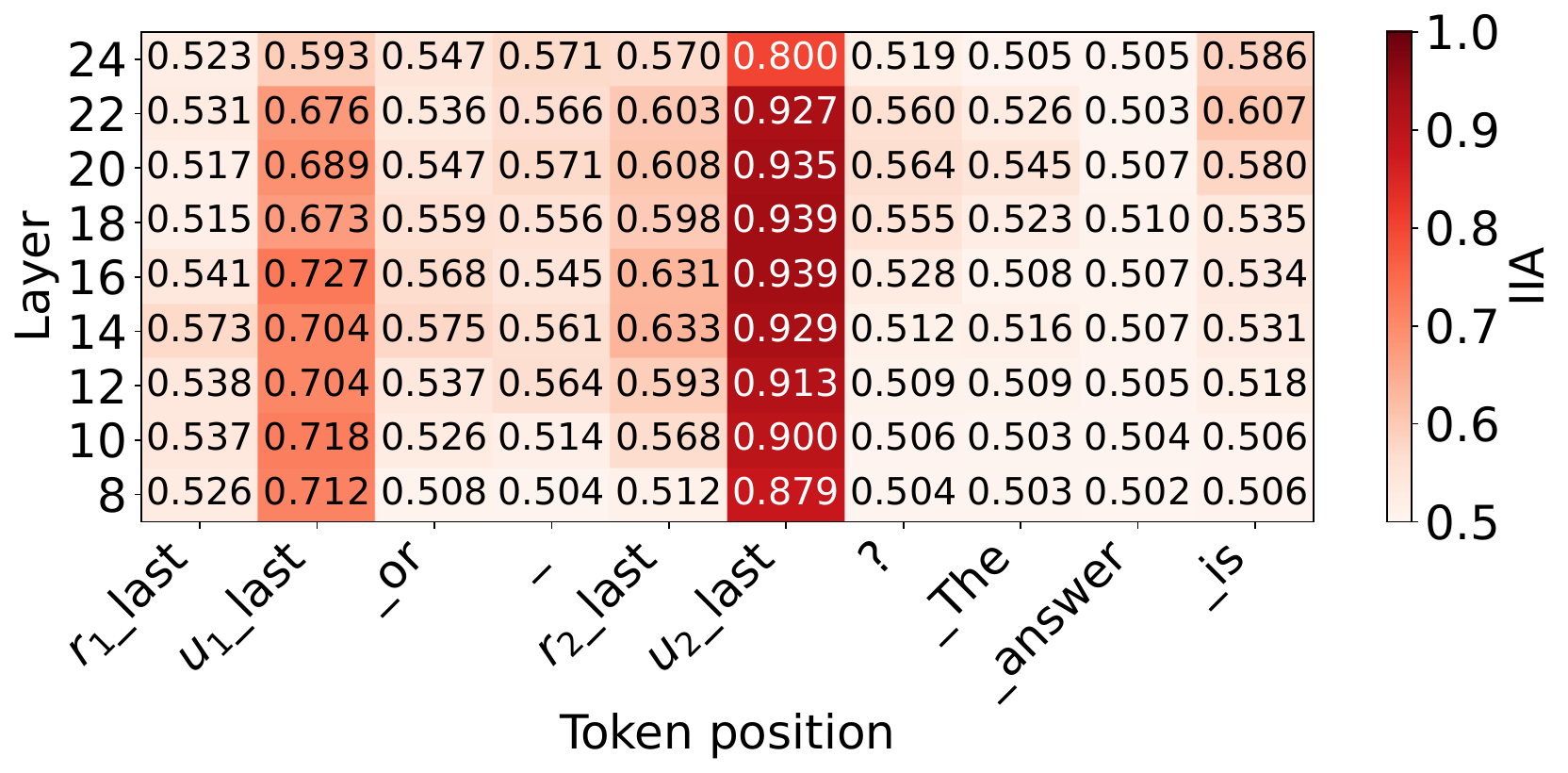}
\caption{
Token-wise DAS intervention accuracy for the \(\mathrm{NumLogDiff}\)/\(\mathrm{UnitLogDiff}\) setting in Qwen3-4B-Base on metric length comparisons.
}
\label{fig:tokenwise_das_numunit}
\end{figure}

\subsection{Experimental Setup}
\label{sec:das_experimental_setup}
We use Distributed Alignment Search (DAS)~\citep{pmlr-v236-geiger24a} to test whether the variables identified by the surrogate analysis are represented in LM activations and causally affect the LM's output.
DAS learns subspaces of the LM representation aligned with high-level variables, and evaluates whether replacing these subspaces with those from source examples changes the LM's output according to the corresponding high-level counterfactual.
We report interchange intervention accuracy (IIA), the fraction of interventions for which the LM's intervened prediction matches the high-level counterfactual label.
Full details of the DAS objective and intervention operation are provided in \cref{sec:appendix_das}.

Our main high-level variables are the numerical difference and unit-scale difference:
\begin{equation}
    Z_1=\mathrm{NumLogDiff}, \quad
    Z_2=\mathrm{UnitLogDiff}.
\end{equation}
For an intervention example \(e=(b,s_1,s_2)\), we jointly intervene on both variables by taking \(\mathrm{NumLogDiff}\) from \(s_1\) and \(\mathrm{UnitLogDiff}\) from \(s_2\).
The high-level counterfactual label is computed as
\begin{equation}
\begin{aligned}
    y_{\mathrm{cf}}^{\mathrm{int}}(e)
    =
    \mathbbm{1}
    [&\mathrm{NumLogDiff}(s_1) \\
     &+\mathrm{UnitLogDiff}(s_2)>0].
\end{aligned}
\end{equation}

\begin{figure}[t!]
\includegraphics[width=\linewidth]{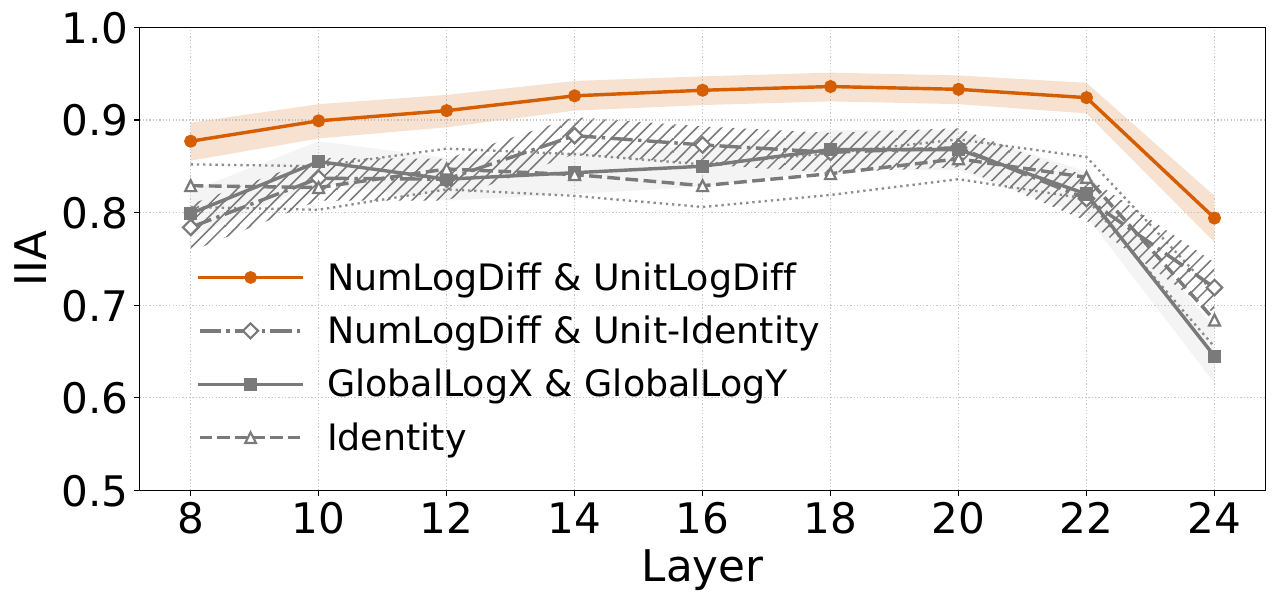}
\centering
\caption{
Layer-wise DAS intervention accuracy at the last token of \(u_2\) for Qwen3-4B-Base on metric length comparisons.
}
\label{fig:das_qwen3-4b_meter}
\end{figure}

We compare this setting against three alternative high-level variable sets.
The first is a shared-scale global-quantity baseline, which tests whether LM activations are better aligned with the two quantities represented on a common scale:
\begin{equation}
    Z_1=\mathrm{GlobalLogX}, \qquad
    Z_2=\mathrm{GlobalLogY}.
\end{equation}
The second is an identity baseline with four variables corresponding to the primitive log-scale numerical and unit-scale components, which tests whether the intervention effect can be explained by the original numerical and unit-scale components rather than by difference variables:
\begin{equation}
\begin{aligned}
    Z_1 &= \log r_1,
    &
    Z_2 &= \log s_D(u_1),\\
    Z_3 &= \log r_2,
    &
    Z_4 &= \log s_D(u_2).
\end{aligned}
\end{equation}

The third is a NumLogDiff \& Unit-Identity baseline, which combines the numerical-difference variable with the original unit-scale identity variables. 
This baseline tests whether the learned subspaces associated with the unit variables in our NumLogDiff \& UnitLogDiff setting are better explained by unit-scale differences than by side-specific lexical signatures.

\begin{equation}
\begin{aligned}
    Z_1 &= \mathrm{NumLogDiff},
    &
    Z_2 &= \log s_D(u_1), \\
    Z_3 &= \log s_D(u_2),
\end{aligned}
\end{equation}

For all DAS settings, we fix the total dimensionality of the non-residual intervention subspaces to \(1024\), allocated evenly across variables.
Thus, the two-variable settings use \(512\) dimensions per variable, the three-variable setting uses \(341\) dimensions per variable (with one remaining dimension unused), while the four-variable identity baseline uses \(256\) dimensions per variable.
As robustness checks, we also vary the total intervention dimensionality to \(256\), \(512\), and \(2048\), evaluate intervention accuracy using 95\% confidence intervals estimated from 10,000 random bootstrap sampling of the evaluation set, and repeat DAS training with three different random seeds.
Across all evaluations, we observe the same qualitative pattern, with details provided in \cref{sec:appendix_das}.
Each intervention example consists of a base input and one source input for each high-level variable.
We use 20K intervention examples for training and 1K held-out examples for evaluation.
We balance output-changing and output-preserving interventions in equal proportions, so the chance-level IIA is \(0.5\).
For all other experimental details, we follow the LMs, prompt templates, and unit settings described in \cref{sec:overall_experimental_setup}.

\subsection{Results}
\label{sec:das_results}

\cref{fig:tokenwise_das_numunit} shows the DAS intervention accuracy across layers and token positions for the \(\mathrm{NumLogDiff}\)/\(\mathrm{UnitLogDiff}\) setting.
The strongest effects concentrate around the last token of \(u_2\).
This suggests that information relevant to the comparison is integrated near the end of the second quantity.
We therefore focus on the last token of \(u_2\) for the layerwise analysis below.

\cref{fig:das_qwen3-4b_meter} reports layerwise DAS results at the last token of \(u_2\) for Qwen3-4B-Base on metric length comparisons.
The \(\mathrm{NumLogDiff}\)/\(\mathrm{UnitLogDiff}\) setting achieves the highest IIA across all reported layers.
This indicates that intervening on the learned subspaces for numerical difference and unit-scale difference can causally steer the LM's comparison output in the direction predicted by the high-level counterfactual.

The \(\mathrm{GlobalLogX}\)/\(\mathrm{GlobalLogY}\), identity, and NumLogDiff \& Unit-Identity baselines achieve above-chance IIA, indicating that global quantities and primitive input components are represented to some extent.
However, all three remain consistently below the \(\mathrm{NumLogDiff}\)/\(\mathrm{UnitLogDiff}\) setting.
These gaps suggest that LM comparison outputs are more effectively steered by interventions aligned with \(\mathrm{NumLogDiff}\)/\(\mathrm{UnitLogDiff}\) variables than by interventions aligned with the tested alternatives, and that the advantage of the UnitLogDiff setting is not fully explained by side-specific unit identities.

Thus, the numerical difference and unit-scale difference identified in the surrogate analysis are not merely correlated with LM behavior; they are represented in a way that can causally affect the LM's decisions.
We observe the same qualitative pattern across other LMs, unit settings, and prompt templates.
Full results are provided in \cref{sec:appendix_das}.

\section{Related Work}
Prior work~\citep{park-etal-2022-language,spokoyny-etal-2022-masked,gopfert-etal-2022-measurement,xu-etal-2024-numcot,10597993,bui-etal-2025-generalization} has examined empirical behavior of LMs on measurement-unit tasks, including basic measurement understanding~\citep{park-etal-2022-language,spokoyny-etal-2022-masked}, measurement extraction from text~\citep{gopfert-etal-2022-measurement}, unit-aware chain-of-thought reasoning~\citep{xu-etal-2024-numcot}, dimension-aware quantitative reasoning~\citep{10597993}, and generalization across measurement systems~\citep{bui-etal-2025-generalization}. 
Building on these behavioral and task-oriented findings, we provide, to our knowledge, the first internal-mechanism analysis of comparing quantities with measurement units.

Prior work has shown that numerical information can be read out from the hidden states of LMs and that this information sometimes causally affects their behavior.
Numerical values and attributes have been analyzed through linear directions~\citep{zhu-etal-2025-language}, low-dimensional subspaces~\citep{heinzerling-inui-2024-monotonic,el-shangiti-etal-2025-geometry,alquabeh2026number}, digit-wise structure~\citep{levy-geva-2025-language}, and helical representations~\citep{kantamneni2025language}. 
For numerical comparison, behavioral work has found magnitude-comparison effects in LMs~\citep{shah-etal-2023-numeric}, and mechanistic studies have analyzed how LMs compute greater-than judgments~\citep{hanna2023how}.
More closely related, \citet{yuchi-etal-2026-llms} show that hidden states encode both the magnitudes of numbers written in different numerical notations, such as standard and exponential notation \(5.7 \times 10^2\), and which number is larger.
While these works focus on numerical values alone, our setting requires LMs to integrate numerals with symbolic unit scales, which introduces an additional compositional step.
In addition, we not only find where the answer is encoded, but also how the numerals and units jointly contribute to the LM's behavior.

Regarding how LMs make judgments through heuristic aggregation, \citet{nikankin-etal-2025-arithmetic} argue that, for the four basic arithmetic operations, LMs solve arithmetic problems by a diverse collection of nontrivial input- and result-pattern heuristics rather than exact algorithms.
In contrast, our results suggest that quantity comparison is governed more centrally by the numerical difference and the unit-scale difference.

\section{Conclusion}
\label{sec:conclusion}
We studied how language models compare quantities with measurement units, such as \textit{110 cm} and \textit{1.2 m}.
Across controlled unit settings, LM accuracy decreases near the comparison boundary.
Through surrogate analysis, we found that LM preferences are better explained by numerical-difference and unit-scale-difference cues than by exact shared-scale quantities.
Through DAS interventions, we further showed that numerical difference and unit-scale difference are represented in activation space and can causally steer LM's outputs.
Overall, our results suggest that LM behavior is better explained by a heuristic account centered on numerical and unit-scale cues than by an account based solely on exact unit conversion.
This provides a controlled case study of how numerical information and symbolic measurement units are combined in language-model representations.

\section{Limitations}
\label{sec:limitations}
While our work shows that language models compare quantities with measurement units using heuristics based on numerical difference and unit-scale difference, it has several limitations.

First, our study focuses on controlled, single-step comparisons between two quantities with measurement units.
This design allows us to analyze LM behavior in a controlled setting, but it does not cover more realistic multi-step quantitative reasoning, where unit conversion may be combined with retrieval, arithmetic, or reasoning over longer textual contexts.
Extending the analysis to such settings is an important direction for future work.

Second, our findings concern the relative explanatory power of the candidate high-level accounts considered in this study.
Our main surrogate analysis uses linear models for interpretability.
Although additional experiments with MLPs and sigmoid-transformed features support our main findings~\cref{sec:nonlinear_surrogate}, our analysis may not capture more complex interactions among cues or heuristic strategies beyond those considered in this study.
Similarly, our DAS analysis evaluates the alignment between candidate high-level causal models and LM behavior under interventions on linear subspaces, rather than fully identifying the underlying computational circuits.
Thus, our analyses neither uniquely identify the computational mechanisms underlying quantity comparison in LMs nor completely rule out alternative explanations.

Third, extending our analysis to reasoning-oriented LMs or long chain-of-thought outputs is nontrivial.
Such LMs may distribute comparison-relevant information over many generated tokens, making it less clear where and when unit-comparison information is encoded and used.
Our current analysis focuses on short answer generation, where the relevant comparison information is localized around the input tokens.
Future work could adapt the proposed analyses to reasoning LMs and multi-token reasoning traces.

\clearpage
\section*{Ethical Considerations}
All data created and/or used in this work was
synthetically generated.
All language models used in this study are publicly available. 
We strictly adhered to the terms and conditions of each model’s license.
During the development of code and the writing of this paper, we made use of AI assistants,
including large language models. 
All code snippets and textual content generated with the assistance of such tools were carefully reviewed and revised by the authors to ensure scientific integrity, accuracy, and ethical compliance.

\section*{Acknowledgments}
This work was supported by JST BOOST
Grant Numbers JPMJBS2412, JPMJBS2421, JPMJBY24D2, JPMJBY24F9; JSPS KAKENHI Grant Number JP25K03175; AMED Grant Number JP25wm0625405, and the Nakajima Foundation.
This work was supported by ABCI 3.0, which is provided by AIST and AIST Solutions.

\bibliography{custom}

\clearpage
\appendix
\section{Task Formulation and Overall Experimental Setup}
\label{sec:appendix_experimental_setup}
This section provides additional details on the task formulation and experimental setup shared across our behavioral, surrogate, and causal analyses.
We first describe the prompt templates and unit notations used in the behavioral analysis, and then explain the prompt and notation choices used in the surrogate and DAS analyses.

\subsection{Prompt Templates}
For the behavioral analysis in \cref{sec:behavioral_analysis}, we evaluate LMs under a broad set of prompt designs.
The prompts vary along two axes.
First, we vary the comparison direction, asking either for the larger quantity or the smaller quantity.
Second, we vary the position of the two quantities relative to the comparison phrase.
In the postposed design, the two quantities appear after the comparison phrase, as in ``Which is larger, \(q_1\) or \(q_2\)?''.
In the proposed design, the two quantities appear before the comparison phrase, as in ``Between \(q_1\) and \(q_2\), which is larger?''.
For mass comparisons, we use ``heavier'' and ``lighter'' instead of ``larger'' and ``smaller''.
The full set of prompt templates is shown in \cref{tab:prompt_templates}.

We also vary the surface form of unit expressions.
For each unit, we test both a short notation, such as ``cm'', and a long notation, such as ``centimeter''.
The unit notations are listed in \cref{tab:unit_notations}.
This allows us to test whether the observed behavioral patterns depend on a particular surface form of measurement units.

Furthermore, to enable analysis of instruction-tuned LMs, we append ``Answer only with one of the two quantities.'' to the prompt templates in \cref{tab:prompt_templates}, ensuring that the LMs output one of the two quantities immediately after the prompt.

\subsection{Choice of Prompt for Main Analyses}
In the behavioral analysis, we observe some variation across prompt templates and unit notations.
In particular, the postposed design generally yields higher accuracy than the preposed design, and the surface form of the unit notation leads to small differences in performance.
However, the main qualitative pattern is stable across these variations.
Across prompt templates and unit notations, accuracy changes systematically with the Quantity Margin, and performance decreases near the comparison boundary.
Based on this observation, the surrogate analysis in \cref{sec:surrogate_analysis} and the DAS analysis in \cref{sec:internal_mechanism} use the first larger-comparison prompt in \cref{tab:prompt_templates}, with the postposed design and short unit notation.
For Qwen3-4B-Base, we also confirm that the second larger-comparison prompt shows the same qualitative pattern.

\subsection{Generation and Evaluation}
All generations are performed with greedy decoding.
For the prompt templates in \cref{tab:prompt_templates}, the evaluated LMs consistently output either \(q_1\) or \(q_2\).
This property allows us to use a unified evaluation protocol across the three analyses.
For the behavioral analysis, we evaluate predictions by exact match against the correct quantity.
For the surrogate analysis, we compare the LM's average log-probability of generating the two candidate answer strings, \(q_1\) and \(q_2\), as continuations of the same prompt.
For the DAS analysis, the same binary output structure allows us to train interventions using the correct candidate string as the supervision signal.

\subsection{Effects of Number Formatting, Unit Notation, and Tokenization}
\label{sec:number_formatting}
First of all, we simply used the default tokenizers provided with each LM.
All values were rendered in ordinary decimal notation; scientific notation was not used. 
The fractional part always had three digits, while the integer part had zero to four digits depending on the value.

In Qwen3, digits and decimal points are mostly tokenized character by character. 
In Olmo3, integer and fractional parts are mainly tokenized in three-digit groups, with the decimal point as a separate token.

We also examined the short and long unit forms in \cref{tab:unit_notations}. 
The long forms corresponding to mm, cm, km, mg, and kg are split into two tokens; most other short and long forms are single tokens. This pattern is shared by Qwen3 and Olmo3.

These surface differences may affect absolute accuracy.
\cref{sec:appendix_behavioral_notationwise} indeed reports a tendency for long notation to outperform short notation.
However, they are unlikely to drive the main findings.
Qwen3 and Olmo3 exhibit consistent qualitative patterns despite differences in number tokenization, and similar results are observed across multiple unit systems and unit notations.
Furthermore, the surrogate analysis uses the average token log-probability of each answer string, reducing (though not necessarily eliminating) the effect of answer-token length on the preference margin, and uses identical prediction targets across surrogate models.

\subsection{Motivation for Analyzing the Quantity Margin in Log Space}
\label{sec:motivation_qm}
Our main empirical observation is that LM accuracy decreases near the comparison decision boundary.
\cref{sec:behavioral_analysis} establishes this phenomenon, \cref{sec: cue-agreement-hypothesis} proposes an explanation based on the cue-combination hypothesis, and \cref{sec:surrogate_analysis} and \cref{sec:internal_mechanism} test this hypothesis.
This line of analysis requires a quantity that measures how close a comparison is to the decision boundary, motivating the introduction of the Quantity Margin.

One motivation for defining the Quantity Margin in log space comes from prior work, which suggests that LMs represent numerical scale logarithmically \citep{alquabeh2026number}.

A second motivation comes from our own preliminary experiments, in which the log-scale Quantity Margin could be decoded substantially more accurately from the hidden state than the corresponding linear-scale margin.
Specifically, in the metric-length setting, we used the hidden state of the last token of \(u_2\) at Layer 18 of Qwen3-4B-Base when prompted with the template defined in \cref{eq:input}.
Using this representation as input, we separately trained PLS (Partial Least Squares) regressions~\citep{WOLD2001109} to predict the log-scale Quantity Margin defined in \cref{eq:quantity_margin} and the corresponding linear-scale margin.
The linear-scale formulation is given by
\begin{equation}
    QM(q_1, q_2)_{\mathrm{linear}}=r_1 s_D(u_1)-r_2s_D(u_2).
\end{equation}

The regressions were trained on 5,000 examples and evaluated on 1,000 held-out examples.
The number of PLS components was varied from 10 to 20.
Performance was evaluated using both \(R^2\) and the mean absolute error (MAE).

\begin{figure*}[t!]
\includegraphics[width=\linewidth]{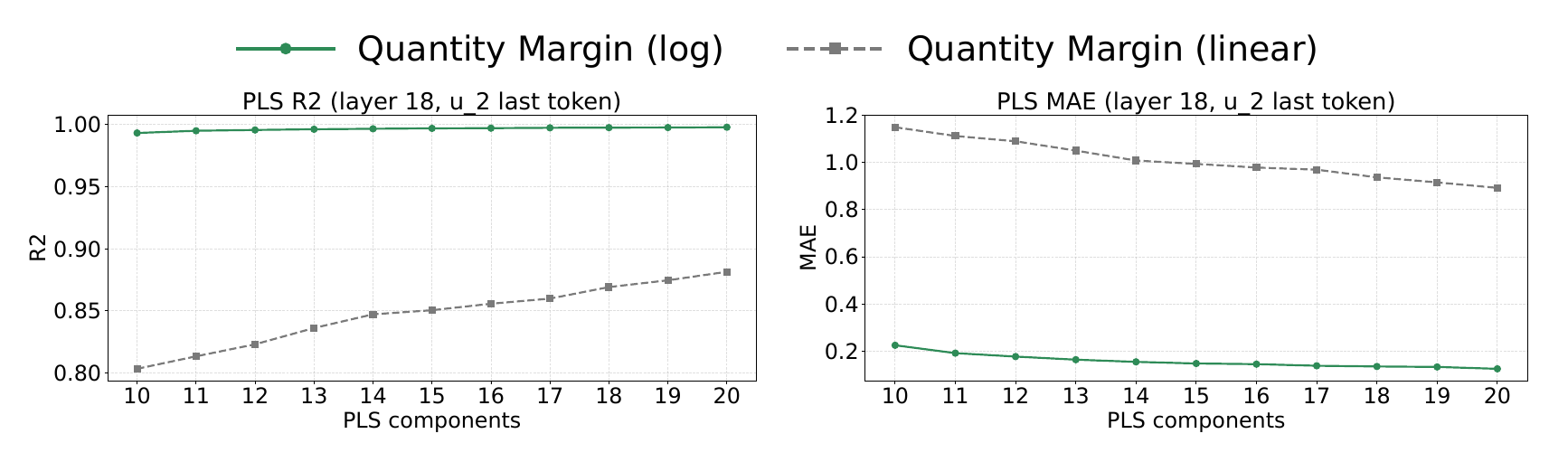}
\centering
\caption{
PLS decoding performance for the log-scale Quantity Margin and the corresponding linear-scale margin from the Layer-18 hidden state at the last token of \(u_2\) in Qwen3-4B-Base (metric-length setting). 
Left: \(R^2\).
Right: mean absolute error (MAE). 
Across all numbers of PLS components, the log-scale Quantity Margin is decoded substantially more accurately than the linear-scale margin, supporting the use of the log-space formulation.
}
\label{fig:pls_results}
\end{figure*}

The results are shown in \cref{fig:pls_results}.
The log-scale Quantity Margin is substantially easier to decode than the corresponding linear-scale margin across all evaluated numbers of PLS components.
These results provide empirical support for our choice to formulate the Quantity Margin in log space.

\section{Heuristics and Heuristic Aggregation}
\label{sec:definition_of_heuristic}

\subsection{Terminology and Formal Definitions}
In this section, we define the terms ``Heuristic,'' ``Heuristic Aggregation,'' and ``Bag of Heuristics'' as used in this paper. 

Our methods (surrogate analysis and DAS) are intended to provide a simpler, more interpretable, high-level account of otherwise opaque LM predictions, following the framework of causal abstraction proposed by \citet{geiger2025ca}. Consequently, neither our methods nor the resulting findings uniquely identify the computational mechanism used by LMs for quantities with measurement units.

One concrete form of such a high-level account is to explain LM behavior through combinations of simple, interpretable heuristics, as in the ``bag of heuristics'' analysis of \citet{nikankin-etal-2025-arithmetic}. Specifically, \citet{nikankin-etal-2025-arithmetic} use the term ``bag of heuristics'' to describe an account that explains LM behavior through a combination of multiple heuristics. 
Our use of the term follows this interpretation.

In this paper, we define a \emph{heuristic} as a comparison cue that is predictive of the LM's answer but for which a comparison rule based on that cue alone does not agree with the correct comparison rule over the full input domain.
For example, the Quantity Margin fully implements the correct comparison rule and is therefore not itself a heuristic. 
We further use the term \emph{heuristic aggregation} to denote a weighted combination of such heuristics. 
We provide the following formal definitions for clarity.

\paragraph{Heuristic.}
For quantities $q_1$ and $q_2$, a heuristic is a comparison rule based on simple, interpretable cues, such as numerical difference, unit-scale difference, unit identity, or thresholds, that does not agree with the correct comparison rule $g$ over the full input domain.
Formally, a comparison rule $f$ is a heuristic if
\begin{equation}
\exists (q_1,q_2)
\quad
f(q_1,q_2)\neq g(q_1,q_2).
\end{equation}
For example, a rule based only on the numerical values or only on the unit scales is a heuristic. 
By contrast, the sign of the Quantity Margin agrees with the correct comparison rule over the full domain and is therefore not a heuristic.

\paragraph{Heuristic Aggregation.}
Let $c_1,\ldots,c_m$ denote heuristic cues. 
Heuristic aggregation is a high-level account that combines these cues using weights $w_j$,
\begin{equation}
    A(q_1,q_2)=\sum_{j=1}^{m}w_j\,c_j(q_1,q_2),
\end{equation}
to explain the LM's preference or output margin. 
This abstraction does not uniquely identify the LM's underlying computational mechanism. 
It explains complex behavior through a weighted combination of simple, interpretable cues.

\subsection{The Role of Heuristics in Our Study}
While our use of the term ``bag of heuristics'' follows that of \citet{nikankin-etal-2025-arithmetic}, our definition of heuristics differs slightly, reflecting the different starting points of the two studies.
\citet{nikankin-etal-2025-arithmetic} identify neurons important for arithmetic tasks and characterize heuristics based on their activation patterns and functional roles.
In contrast, we define heuristics and the correct comparison rule based on the structure of the quantity comparison task, independently of the internal mechanisms of a particular LM.

This distinction also reflects a difference in research objectives.
\citet{nikankin-etal-2025-arithmetic} aim to understand the computational mechanisms underlying arithmetic reasoning in LMs, particularly whether LMs implement generalizable algorithms or rely on memorization.
Our aim, in contrast, is to evaluate which of several candidate high-level explanations best captures LM behavior.
Surrogate analysis evaluates how well each candidate explanation predicts LM behavior, while DAS evaluates how well the corresponding high-level causal model aligns with LM behavior under interventions.
A particular strength of our task-driven approach is that it enables systematic and quantitative comparisons among alternative explanations, including those based on the correct comparison rule.
At the same time, our methods do not uniquely identify the LM's computational algorithm or completely rule out alternative explanations, including the possibility of exact computation.

Furthermore, heuristic-based and exact computations need not be mutually exclusive and may coexist within the same LM.
Our finding that a heuristic aggregation account better captures LM behavior than alternative accounts concerns their relative explanatory power as high-level abstractions.

\section{Behavioral Observation: LMs Become Less Accurate Near the Boundary}
\label{sec:appendix_behavioral}
In \cref{sec:behavioral_analysis}, we showed that LM accuracy in quantity comparison is strongly organized by the Quantity Margin.
Accuracy remains high when the absolute margin is large, but decreases as the comparison approaches the decision boundary.
In this section, we examine the robustness of this gradual margin-dependent pattern and the additional differences that arise across LMs, prompt templates, and unit notations.

\subsection{Results Across LMs}
\label{sec:appendix_behavioral_modelwise}
\begin{figure*}[t!]
\includegraphics[width=\linewidth]{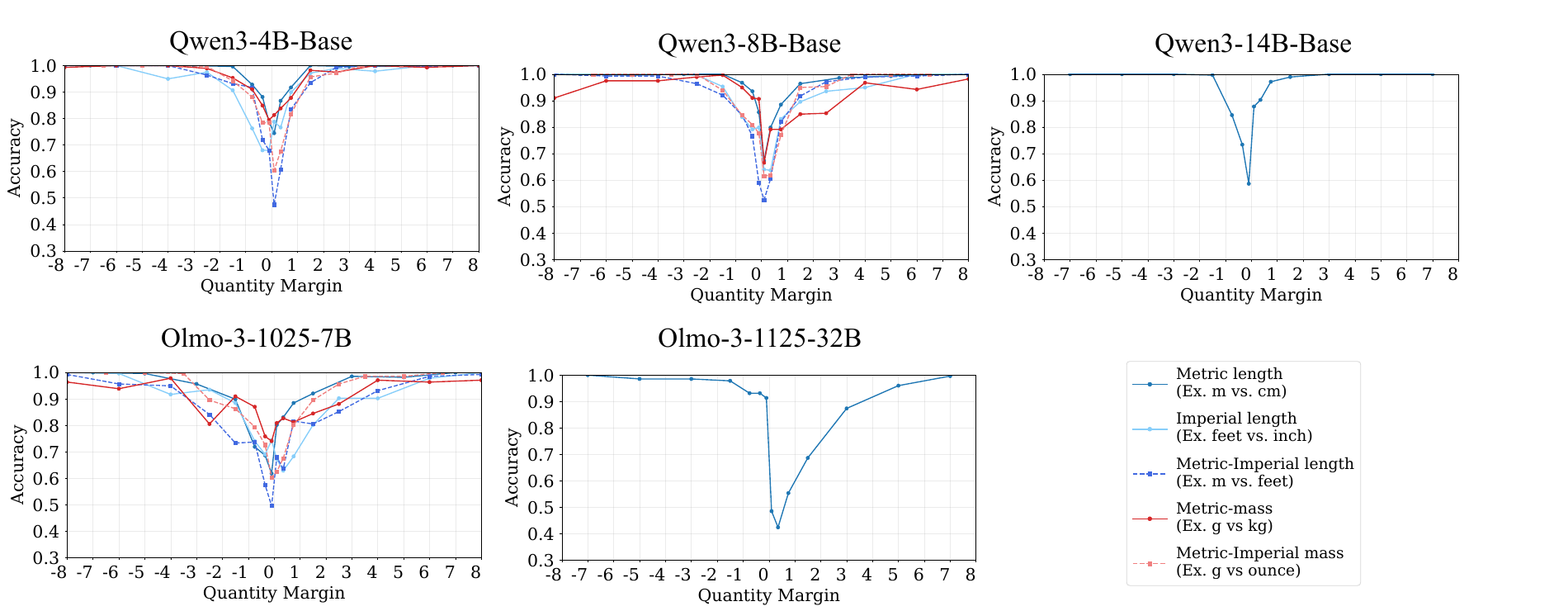}
\centering
\caption{
Accuracy of comparisons between quantities with measurement units, grouped by Quantity Margin for Qwen3-4B-Base, Qwen3-8B-Base, Qwen3-14B-Base, Olmo-3-1025-7B, and Olmo-3-1125-32B.
Blue and red denote length and mass comparisons, respectively.
Dashed curves denote heterogeneous Metric-imperial comparisons.
}
\label{fig:behavioral_analysis_globaldiff_modelwise}
\end{figure*}

\begin{figure*}[t!]
\includegraphics[width=\linewidth]{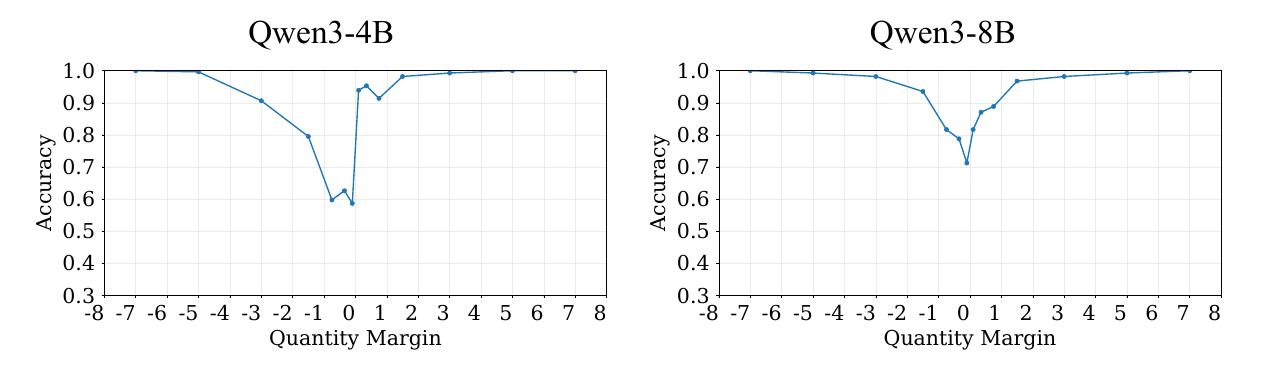}
\centering
\caption{
Accuracy of comparisons between quantities with measurement units, grouped by Quantity Margin for the instruct LMs (Qwen3-4B, Qwen3-8B) in metric length setting.
}
\label{fig:behavioral_analysis_instructmodels}
\end{figure*}

\cref{fig:behavioral_analysis_globaldiff_modelwise} shows the results for Qwen3-4B-Base, Qwen3-8B-Base, Qwen3-14B-Base, Olmo-3-1025-7B, and Olmo-3-1125-32B.
In addition, \cref{fig:behavioral_analysis_instructmodels} reports the corresponding results for the instruct LMs Qwen3-4B and Qwen3-8B.

Across all five base LMs and two instruction-tuned LMs, accuracy changes systematically with the Quantity Margin.
Comparisons are easy when the absolute margin is large and become gradually more difficult as the margin approaches zero.
This indicates that the boundary-related degradation observed in \cref{sec:behavioral_analysis} is not specific to a single LM.

\cref{fig:behavioral_analysis_globaldiff_modelwise} also shows differences across unit settings.
Heterogeneous metric-imperial comparisons tend to be less accurate than comparisons within a single unit system.
For length comparisons, metric comparisons are generally easier than imperial comparisons.
Overall, the same margin-dependent pattern is preserved across LMs and unit settings.

\subsection{Effects of Prompt Templates}
\label{sec:appendix_behavioral_promptwise}
\begin{figure*}[t!]
\includegraphics[width=\linewidth]{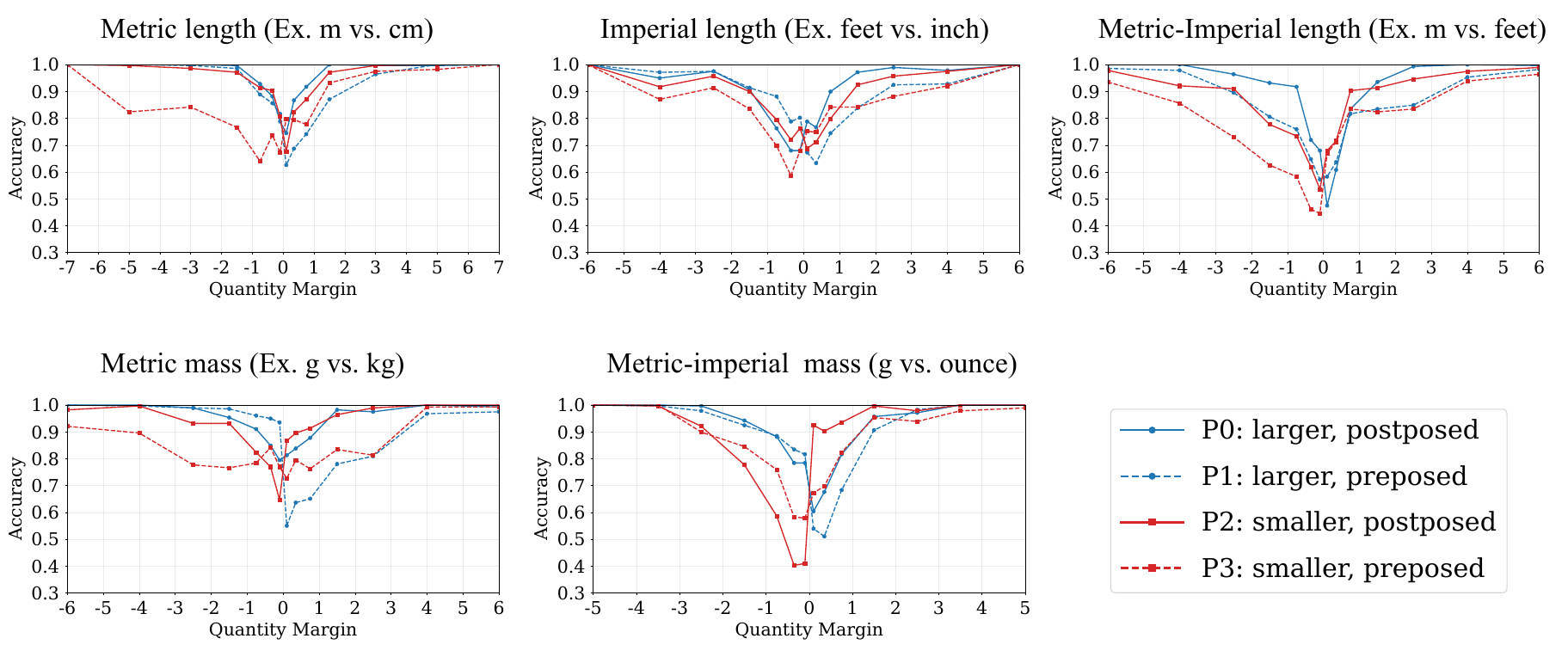}
\centering
\caption{
Accuracy of comparisons between quantities with measurement units, grouped by Quantity Margin for Qwen3-4B-Base.
Each panel corresponds to a unit setting.
Curves compare four prompt designs: larger-postposed, larger-preposed, smaller-postposed, and smaller-preposed.
}
\label{fig:behavioral_analysis_globaldiff_promptwise}
\end{figure*}

We next examine the effect of prompt design.
As described in \cref{tab:prompt_templates}, we vary both the comparison direction and the position of the quantities relative to the comparison phrase.
For the comparison direction, we ask either for the larger quantity or for the smaller quantity.
For the phrase position, we compare a postposed design, where the quantities \(q_1\) and \(q_2\) appear after the comparison phrase, with a preposed design, where the quantities appear before the comparison phrase.
For example, the postposed design uses prompts such as ``Which is larger, \(q_1\) or \(q_2\)?'', whereas the preposed design uses prompts such as ``Between \(q_1\) and \(q_2\), which is larger?''.
\cref{fig:behavioral_analysis_globaldiff_promptwise} shows the results across these prompt templates for Qwen3-4B-Base.
The main margin-dependent pattern is stable across prompt designs.
For all prompt templates, comparisons become gradually more difficult as the Quantity Margin approaches zero.
At the same time, the prompt wording affects the absolute accuracy.
Although the trend is not perfectly consistent across all unit settings, the postposed prompts, shown by solid curves, tend to achieve higher accuracy than the preposed prompts, shown by dashed curves.
Similarly, prompts asking for the larger quantity tend to perform better than prompts asking for the smaller quantity, although this effect is also not fully consistent across all settings.

\subsection{Effects of Unit Notation}
\label{sec:appendix_behavioral_notationwise}
\begin{figure*}[t!]
\includegraphics[width=\linewidth]{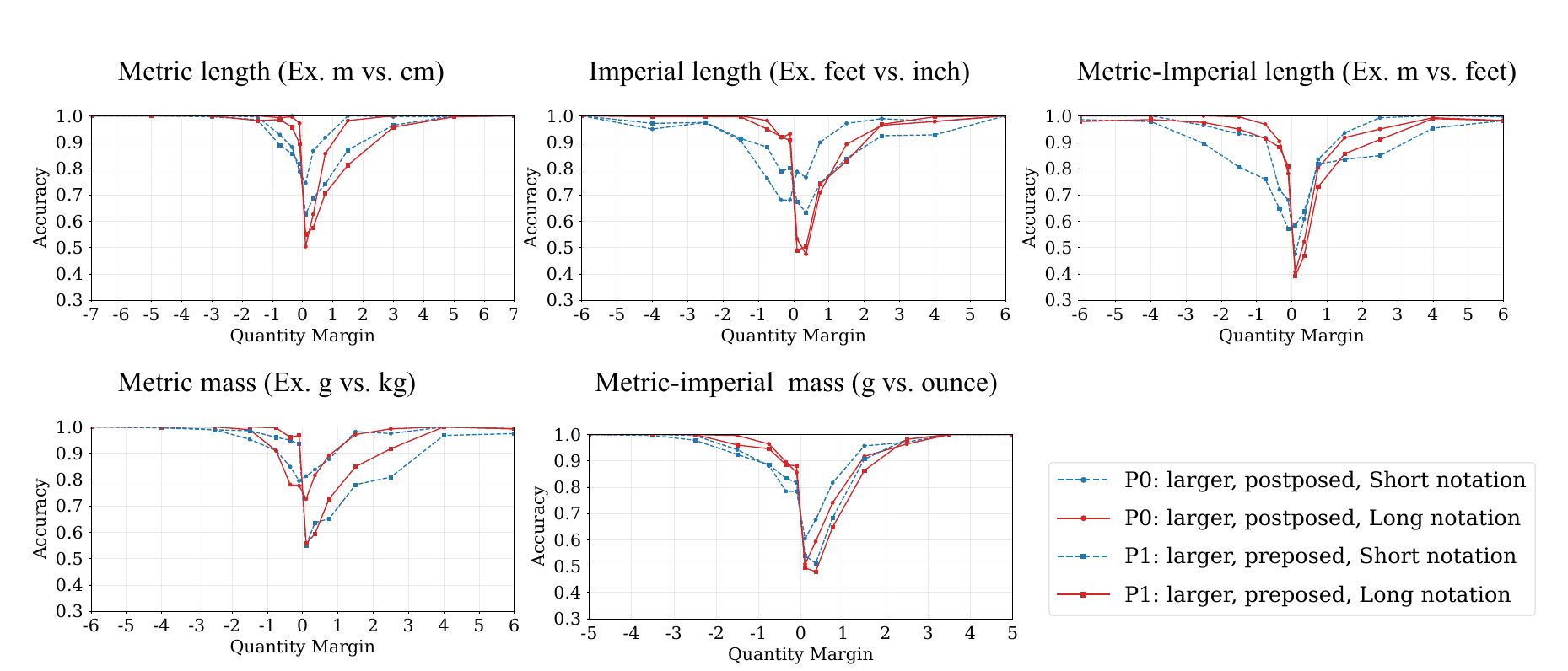}
\centering
\caption{
Accuracy of comparisons between quantities with measurement units, grouped by Quantity Margin for Qwen3-4B-Base.
Each panel corresponds to a unit setting.
Curves compare short-unit notation, such as \textit{cm}, with long-unit notation, such as \textit{centimeter}, across prompt designs.
}
\label{fig:behavioral_analysis_globaldiff_notationwise}
\end{figure*}

We also examine whether the surface form of units affects LM behavior.
As described in \cref{tab:unit_notations}, we compare short unit notation, such as \textit{cm}, with long unit notation, such as \textit{centimeter}.
This allows us to test whether the observed behavioral patterns depend on the surface form used to express measurement units.
\cref{fig:behavioral_analysis_globaldiff_notationwise} shows the results for Qwen3-4B-Base.
The margin-dependent pattern again remains stable across unit notations.
For both short and long unit forms, accuracy decreases as the Quantity Margin approaches zero.
However, the choice of notation affects the absolute accuracy.
Although the trend is not perfectly consistent across all settings, long-unit notation, shown by red curves, tends to achieve higher accuracy than short-unit notation, shown by blue curves.
These results suggest that unit surface forms can influence performance, while the overall boundary-related degradation remains robust.

\subsection{Additional Behavioral Controls}
\label{sec:additional_behavioral_controls}

We conduct additional experiments to examine the effects of number formatting, input order, and answer format on LM behavior.

\paragraph{Number format ablation.}
As described in \cref{sec:number_formatting}, we render numerical values with three digits after the decimal point.
To examine whether LM behavior depends on this formatting choice, we remove trailing zeros from numerical expressions (e.g., 1.230 becomes 1.23).
Of the 5,000 test examples, 840 examples are affected by this modification.
We find that removing trailing zeros changes the LM's comparison decision in only 1.7\% of these examples, suggesting that LM behavior is largely robust to this change in number formatting.

\paragraph{Anti-symmetry check.}
To examine the effect of input order, we swap $q_1$ and $q_2$ for each of the 5,000 test examples and evaluate Qwen3-4B-Base.
The LM selects the same quantity before and after swapping in 93.90\% of the examples.
This strict per-example check suggests that the effect of position bias is limited.
Furthermore, the order of the two quantities is randomized during data generation, further mitigating the effect of position bias on our evaluation.

\paragraph{A/B choice labels.}
To examine whether the observed behavioral patterns depend on the answer format, we additionally evaluate LM behavior using A/B choice labels instead of full answer strings.
Since the choice labels themselves may introduce bias, we evaluate each example twice, first assigning A to $q_1$ and B to $q_2$, and then reversing the assignment.
For each assignment, we compute the difference in log-probability between the label corresponding to $q_1$ and that corresponding to $q_2$.
We average these two preference margins and select $q_1$ if the averaged margin is positive and $q_2$ if it is negative.
\cref{tab:behavioral_answer_format} shows the results.
Although the A/B choice format yields slightly higher accuracy, both answer formats exhibit a consistent decrease in accuracy near the comparison boundary.
These results suggest that the observed pattern is not specific to the full-string answer format.

\subsection{Generated Outputs from Reasoning LMs}
We conducted a behavioral analysis of representative generated outputs from Qwen3-4B-Thinking-2507~\citep{yang2025qwen3technicalreport}.
Representative outputs are provided in \cref{tab:thinking_examples}.
Analysis of these outputs shows that, at the output level, the model often performs correct unit conversion explicitly.
It sometimes considers multiple solution paths, including conversions in both directions.
However, for easy comparisons with a large Quantity Margin, the model may still rely on heuristic statements, such as noting that one quantity is sufficiently larger than the other.

\section{Cue-Combination Hypothesis: Quantity Decisions Depend on Cue Agreement}
\label{sec:appendix_hypothesis}
In \cref{sec: cue-agreement-hypothesis}, we introduced the cue-combination hypothesis, where quantity decisions are explained by the agreement and conflict among comparison cues.
As a concrete example, we analyzed the interaction between numerical-difference and unit-scale-difference cues, \(\mathrm{NumLogDiff}\) and \(\mathrm{UnitLogDiff}\).
In this section, we then demonstrate that the cue-agreement pattern shown in \cref{fig:behavioral_heatmap_numunit_qwen3_4b} is consistently observed across different unit settings.

\begin{figure*}[t!]
\includegraphics[width=\linewidth]{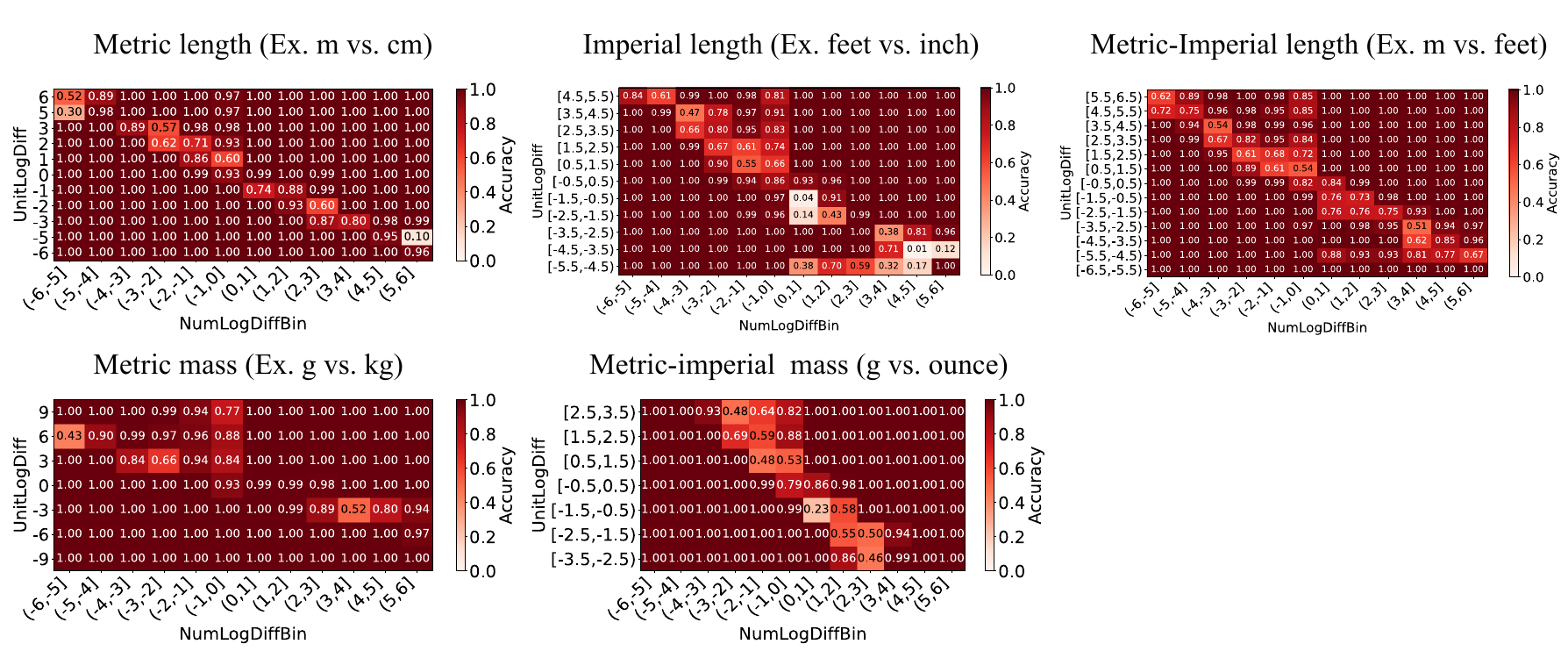}
\centering
\caption{
Accuracy of comparisons between quantities
with measurement units, grouped by NumLogDiff and
UnitLogDiff (Qwen3-4B-Base).
}
\label{fig:behavioral_heatmap_numunit_qwen3_4b_unitwise}
\end{figure*}

\cref{fig:behavioral_heatmap_numunit_qwen3_4b_unitwise} shows LM accuracy grouped by \(\mathrm{NumLogDiff}\) and \(\mathrm{UnitLogDiff}\) for each unit setting.
Across unit settings, accuracy is high in regions where the two cues support the same answer.
In contrast, errors concentrate near the diagonal region where the two cues cancel each other out, yielding a small Quantity Margin.
This shows that the cue-conflict pattern discussed in \cref{sec: cue-agreement-hypothesis} is not specific to metric length comparisons, but is shared across length, mass, and metric-imperial comparisons.

\section{Surrogate Analysis: Number and Unit Differences Best Predict LM Behavior}
\label{sec:appendix_surrogate_analysis}
In \cref{sec:surrogate_analysis}, we showed that LM comparison behavior is well explained by heuristic cues based on numerical difference and unit-scale difference.
In this section, we first describe the full design of the surrogate features.
We then provide additional surrogate results and show that the main trend is consistent across unit settings, LMs, and prompt templates.
In particular, surrogate models based on signed or thresholded \(\mathrm{NumLogDiff}\) and \(\mathrm{UnitLogDiff}\) consistently explain LM behavior well.

\subsection{Surrogate Feature Design}
\label{sec:appendix_surrogate_features}
\cref{tab:heuristic_features} summarizes the full set of heuristic features considered in our surrogate analysis.
We design these features to cover semantically distinct ways of decomposing a quantity comparison while avoiding redundant feature families.

The Quantity Margin in \cref{eq:quantity_margin} is determined by four primitive variables:
the left numeral \(r_1\), the left unit scale \(s_D(u_1)\), the right numeral \(r_2\), and the right unit scale \(s_D(u_2)\).
We therefore begin with features corresponding to these primitive components.
In \cref{tab:heuristic_features}, these include Left number log, Right number log, Left unit-scale log, and Right unit-scale log.

We then construct additional feature families by combining subsets of these four primitive variables.
One possible decomposition combines three variables on one side and one variable on the other.
For example, Left value in right unit together with Right number log represents a comparison between the right numeral and the left quantity converted into the right unit scale.
Similarly, the right value in the left unit together with the left number log represents the corresponding comparison in the left unit scale.

Another decomposition splits the four primitive variables into two pairs.
The split \((r_1,r_2)\) and \((s_D(u_1),s_D(u_2))\) yields the numerical-difference and unit-scale-difference variables, \(\mathrm{NumLogDiff}\) and \(\mathrm{UnitLogDiff}\).
This feature family tests whether LM behavior is explained by separately comparing numerals and unit scales.
The split \((r_1,s_D(u_1))\) and \((r_2,s_D(u_2))\) yield \(\mathrm{GlobalLogX}\) and \(\mathrm{GlobalLogY}\), corresponding to the two quantities represented on a common scale.
Finally, combining all four variables gives the exact Quantity Margin itself, which we include as the Quantity Margin feature in \cref{tab:heuristic_features}.
This organization gives a systematic set of semantically distinct feature families, ranging from primitive input components to difference variables, shared-scale quantities, and the exact comparison rule.

To capture heuristic cues at multiple levels of granularity, we instantiate these quantities in several forms.
Signed features represent the coarsest cues, such as whether a quantity or difference is positive.
Threshold features represent coarse magnitude information, such as whether a log-scale variable exceeds a threshold.
Continuous features use the original log-scale values directly.

After all, the resulting list of individual features is shown in \cref{tab:heuristic_features}.
The surrogate models used in the experiments are then constructed from combinations of these feature families.
The full list of surrogate feature sets is provided in \cref{tab:surrogate_feature_sets}.
Among these, \cref{tab:surrogate_representative_models} selects the representative surrogate families that are most important for interpreting the results, and these are used in the main paper.

\subsection{Capacity-Matched Surrogate Analysis}
\label{sec:nonlinear_surrogate}

To examine whether differences in surrogate capacity account for the observed differences in explanatory power across feature sets, we conduct additional experiments using MLPs and sigmoid-transformed features.
For each feature set, we select hyperparameters on the validation set using the same search space.
We conduct these experiments on Qwen3-4B-Base in the metric length setting.

\paragraph{MLP-based surrogates.}
We use MLPs with a single hidden layer.
For each feature set, we select the hidden width from ${4, 8, 16, 32, 64}$ and the L2 regularization coefficient from ${0, 10^{-5}, 10^{-4}, 10^{-3}, 10^{-2}}$ based on the validation $R^2$.
We train the MLPs using Adam with a learning rate of $10^{-3}$ and a batch size of 256 for up to 2,000 epochs.
We employ early stopping with a patience of 50 epochs and retain the checkpoint with the highest validation $R^2$.

\paragraph{Surrogates with sigmoid-transformed features.}
We transform each feature using the sigmoid function
\begin{equation}
s_k(x) = \frac{1}{1+\exp(-kx)},
\end{equation}
where $k$ controls the steepness of the transformation.
For each feature set, we select $k$ from ${0.01, 0.05, 0.1, 0.5, 1, 5, 10, 50, 100}$ based on the validation $R^2$.

\paragraph{Results.}
\cref{tab:nonlinear_surrogate_results} shows the results.
With both types of surrogates, NumLogDiff and UnitLogDiff jointly achieve substantially higher $R^2$ than the other feature sets on Boundary examples.
Although Quantity Margin achieves the highest $R^2$ on All examples with sigmoid-transformed features, its $R^2$ remains low on Boundary examples.
These results suggest that the higher explanatory power of numerical-difference and unit-scale-difference cues on Boundary examples cannot be attributed solely to differences in surrogate capacity.

\subsection{Surrogate Analysis with Digit-Based Features}
\label{sec:digit_based_surrogate}

To examine whether simpler cues based on the surface form of numerical expressions can account for LM behavior, we additionally consider two digit-based features: FirstDigitDiff and DigitCountDiff.

\paragraph{FirstDigitDiff.}
FirstDigitDiff is defined as the difference between the first digits of the numerical expressions of the two quantities.
For example, for the numerical expressions 12.300 and 9.5, FirstDigitDiff is $1-9=-8$.
We treat a leading zero as the first digit (e.g., the first digit of 0.045 is 0).
This feature captures a comparison cue based on the first digits rather than the numerical values themselves.

\paragraph{DigitCountDiff.}
DigitCountDiff is defined as the difference between the total numbers of digits in the numerical expressions of the two quantities.
For example, for 12.300 and 9.5, DigitCountDiff is $5-2=3$.
This feature captures a comparison cue based on the number of digits in the numerical expressions.

\paragraph{Results.}
\cref{tab:digit_control_surrogate_results} shows the results.
The surrogate model using NumLogDiff and UnitLogDiff thresholds achieves higher $R^2$ than those using digit-based features on both All examples and Boundary examples.
The difference is particularly pronounced on Boundary examples, where the former achieves an $R^2$ of 0.749, while the latter achieve at most 0.186.
These results suggest that the explanatory power of numerical-difference cues cannot be sufficiently accounted for by the digit-based cues tested here.

\subsection{Additional Surrogate Results}
\label{sec:appendix_surrogate_additional_results}
\begin{figure*}[t!]
\includegraphics[width=\linewidth]{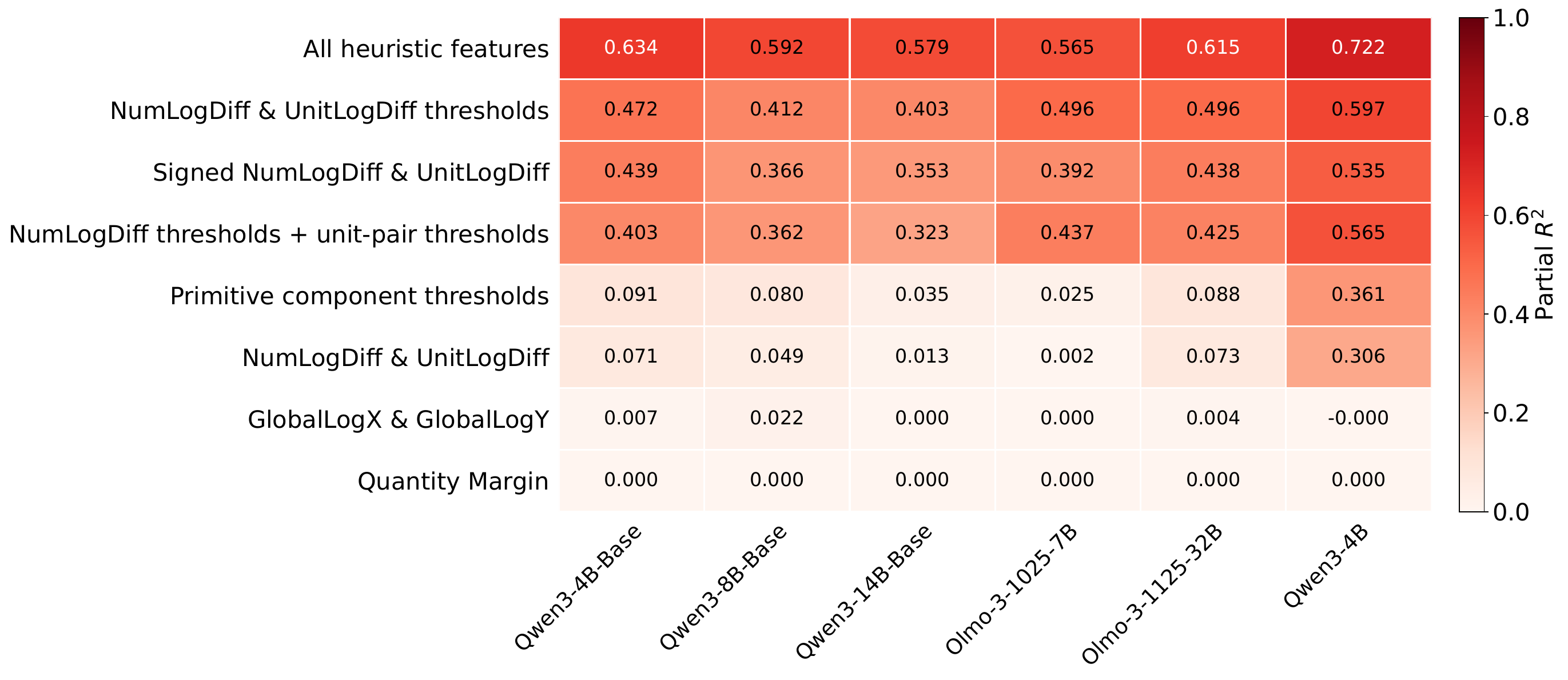}
\centering
\caption{
Surrogate analysis using \(R^2_{\mathrm{partial}}\) in the metric-length setting of the all examples setting. 
A higher \(R^2_{\mathrm{partial}}\) indicates that the surrogate model based on the corresponding features provides a better explanation of the LM's behavior. 
We evaluate five base LMs (Qwen3-4B-Base, Qwen3-8B-Base, Qwen3-14B-Base, Olmo-3-1025-7B, and Olmo-3-1125-32B) as well as the instruction-tuned LM Qwen3-4B.
}
\label{fig:surrogate_predictivity_heatmap_modelwise}
\end{figure*}

\begin{figure*}[t!]
\includegraphics[width=\linewidth]{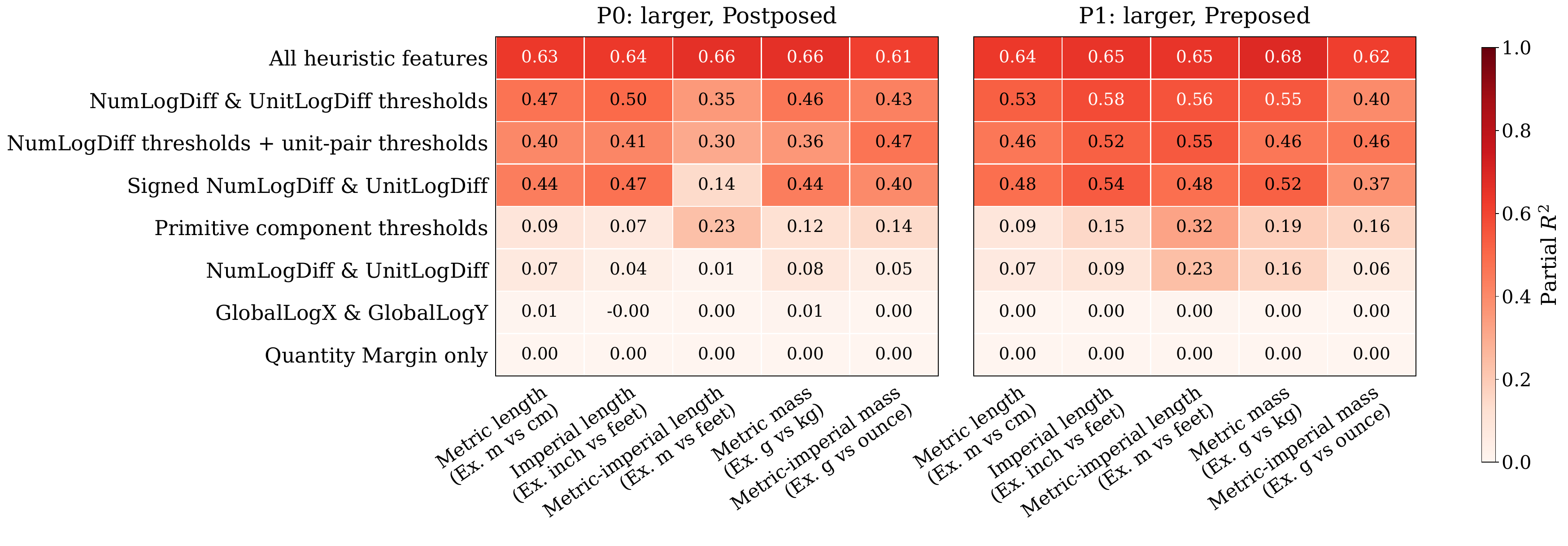}
\centering
\caption{
Surrogate analysis using \(R^2_{\mathrm{partial}}\) across all unit settings in \cref{tab:unit_settings} for Qwen3-4B-Base under different prompt configurations. In the postposed prompt, the quantity expressions appear after the comparison expression ``larger'', whereas in the preposed prompt, they appear before ``larger''.
}
\label{fig:surrogate_predictivity_heatmap_promptwise}
\end{figure*}

We next examine whether the main surrogate-analysis finding holds beyond the representative metric length result in \cref{tab:meter_AbsGlobalDiffBin_hard_abs_lt_02_surrogate_representative_results}.
Using the \(R^2_{\mathrm{partial}}\) defined in \cref{sec:surrogate_analysis}, \cref{fig:surrogate_predictivity_heatmap_modelwise} summarizes the representative surrogate results across LMs.
\cref{fig:surrogate_predictivity_heatmap_promptwise} shows the corresponding comparison across prompt templates and unit settings.

Across these settings, the same qualitative trend appears.
Surrogate models based on numerical difference and unit-scale difference, especially signed or thresholded versions of \(\mathrm{NumLogDiff}\) and \(\mathrm{UnitLogDiff}\), consistently obtain high \(R^2_{\mathrm{partial}}\).
This indicates that the LM log-probability margin is well captured by coarse cues about which side has the larger numeral and which side has the larger unit scale.
By contrast, feature sets based on shared-scale quantities, \(\mathrm{GlobalLogX}\) and \(\mathrm{GlobalLogY}\), tend to obtain lower \(R^2_{\mathrm{partial}}\).

Also, we therefore provide representative surrogate tables for each unit setting:
\cref{tab:meter_AbsGlobalDiffBin_hard_abs_lt_02_surrogate_representative_results} for metric length,
\cref{tab:feet_test_quantity_margin_abs_lt_0_2_r2_representative} for imperial length,
\cref{tab:meter_feet_test_quantity_margin_abs_lt_0_2_r2_representative} for metric-imperial length,
\cref{tab:gram_test_quantity_margin_abs_lt_0_2_r2_representative} for metric mass, and
\cref{tab:gram_pound_test_quantity_margin_abs_lt_0_2_r2_representative} for metric-imperial mass.
These tables show that signed and thresholded \(\mathrm{NumLogDiff}\) and \(\mathrm{UnitLogDiff}\) features explain LM behavior well across unit settings.

We also provide the full detailed surrogate results for the metric length setting.
While \cref{tab:meter_AbsGlobalDiffBin_hard_abs_lt_02_surrogate_representative_results} reports only the representative surrogate families from \cref{tab:surrogate_representative_models}, \cref{tab:meter_surrogate_detailed_results} reports results for all surrogate models defined in \cref{tab:surrogate_feature_sets}.
The raw metric length values are shown in \cref{tab:meter_surrogate_detailed_results}.
Overall, these additional results support the same conclusion as the main analysis: LM quantity-comparison behavior is better explained by numerical-difference and unit-scale-difference heuristics than by exact shared-scale quantities alone.

\section{Causal Analysis: Number and Unit Differences Steer LM Decisions}
\label{sec:appendix_das}
In \cref{sec:internal_mechanism}, we used Distributed Alignment Search (DAS)~\citep{pmlr-v236-geiger24a} to test whether the variables identified by the surrogate analysis are represented in LM activations and causally affect the LM's output.
The main result is that interventions on subspaces aligned with \(\mathrm{NumLogDiff}\) and \(\mathrm{UnitLogDiff}\) steer the LM's comparison decisions more strongly than interventions on the \(\mathrm{GlobalLogX}/\mathrm{GlobalLogY}\) baseline or the identity baseline.
This suggests that numerical difference and unit-scale difference are not merely correlated with LM behavior, but are represented in a way that can causally affect the LM's decisions.
In this section, we provide the formal details of DAS and report additional robustness results across LMs, unit settings, prompt templates, and intervention dimensionalities.

\subsection{Distributed Alignment Search}
We use Distributed Alignment Search (DAS) to test whether the decomposed cues identified by the surrogate analysis are represented in the LMs and causally affect their output.
DAS learns an alignment between high-level causal variables and linear subspaces of a LM's representation by optimizing an interchange intervention objective.

Let \(\mathbf{h}(b) \in \mathbb{R}^d\) be the hidden representation of a base input \(b\), and let \(\mathbf{h}(s_j) \in \mathbb{R}^d\) be the hidden representation of a source input \(s_j\).
For a set of high-level variables \({Z_1},\ldots,Z_k\), DAS learns an orthogonal rotation matrix \(\mathbf{R} \in \mathbb{R}^{d \times d}\) that decomposes the rotated representation space into orthogonal subspaces:
\begin{equation}
    \mathcal{Y}
    =
    \mathcal{Y}_0
    \oplus
    \mathcal{Y}_1
    \oplus
    \cdots
    \oplus
    \mathcal{Y}_k .
\end{equation}
Here, \(\mathcal{Y}_j\) is aligned with the high-level variable \(Z_j\), and \(\mathcal{Y}_0\) is the residual subspace.

A distributed interchange intervention replaces the subspace corresponding to each \(Z_j\) in the base representation with the corresponding subspace from the
source representation:
\begin{align}
    \mathbf{h}^{*}(b)
    =
    \mathbf{R}^{-1}
    \bigg(
        &\mathrm{Proj}_{\mathcal{Y}_0}(\mathbf{R}\mathbf{h}(b))
        \notag \\
        &+
        \sum_{j=1}^{k}
        \mathrm{Proj}_{\mathcal{Y}_j}(\mathbf{R}\mathbf{h}(s_j))
    \bigg).
    \label{eq:das_intervention}
\end{align}
The intervened representation \(\mathbf{h}^{*}(b)\) is then passed through the
remaining LM's layers to obtain the LM's counterfactual output.

For compactness, let \(e=(b,s_1,\ldots,s_k)\) denote an intervention example.
Across DAS settings, base and source inputs are sampled from the same data-generating domain of quantity pairs $(q_1,q_2)$.
We train \(\mathbf{R}\) so that the LM's counterfactual output matches the counterfactual label predicted after intervening on the corresponding high-level variables.
For each intervention example \(e\), we construct a high-level counterfactual by taking the base input and replacing each intermediate variable \(Z_j\) with the value computed from the corresponding source input \(s_j\).
Let \(y_{\mathrm{cf}}^{\mathrm{int}}(e)\) denote this high-level counterfactual label, and let \(\mathbf{p}_{\theta}^{\mathrm{int}}(e)\) denote the LM's output distribution after applying the distributed interchange intervention in \cref{eq:das_intervention}.
The DAS objective minimizes the cross-entropy loss:
\begin{equation}
    \mathcal{L}_{\mathrm{DAS}}
    =
    \mathrm{CE}
    \left(
        \mathbf{p}_{\theta}^{\mathrm{int}}(e),
        y_{\mathrm{cf}}^{\mathrm{int}}(e)
    \right).
    \label{eq:das_objective}
\end{equation}

We evaluate the learned alignment using interchange intervention accuracy (IIA), the fraction of intervention examples for which the LM's intervened prediction matches the high-level counterfactual label:
\begin{equation}
    \mathrm{IIA}
    =
    \frac{1}{|\mathcal{D}_{\mathrm{int}}|}
    \sum_{e\in \mathcal{D}_{\mathrm{int}}}
    \mathbbm{1}
    \left[
        \hat{y}_{\theta}^{\mathrm{int}}(e)
        =
        y_{\mathrm{cf}}^{\mathrm{int}}(e)
    \right],
    \label{eq:iia}
\end{equation}
where \(\hat{y}_{\theta}^{\mathrm{int}}(e)\) is the LM's predicted label after intervention.
A high IIA indicates that intervening on the learned subspaces changes the LM's output consistently with interventions on the corresponding high-level variables.

\subsection{Robustness Across LMs, Unit Settings, Prompts, and training random seeds}
\label{sec:appendix_das_robustness}
\begin{figure*}[t!]
\includegraphics[width=\linewidth]{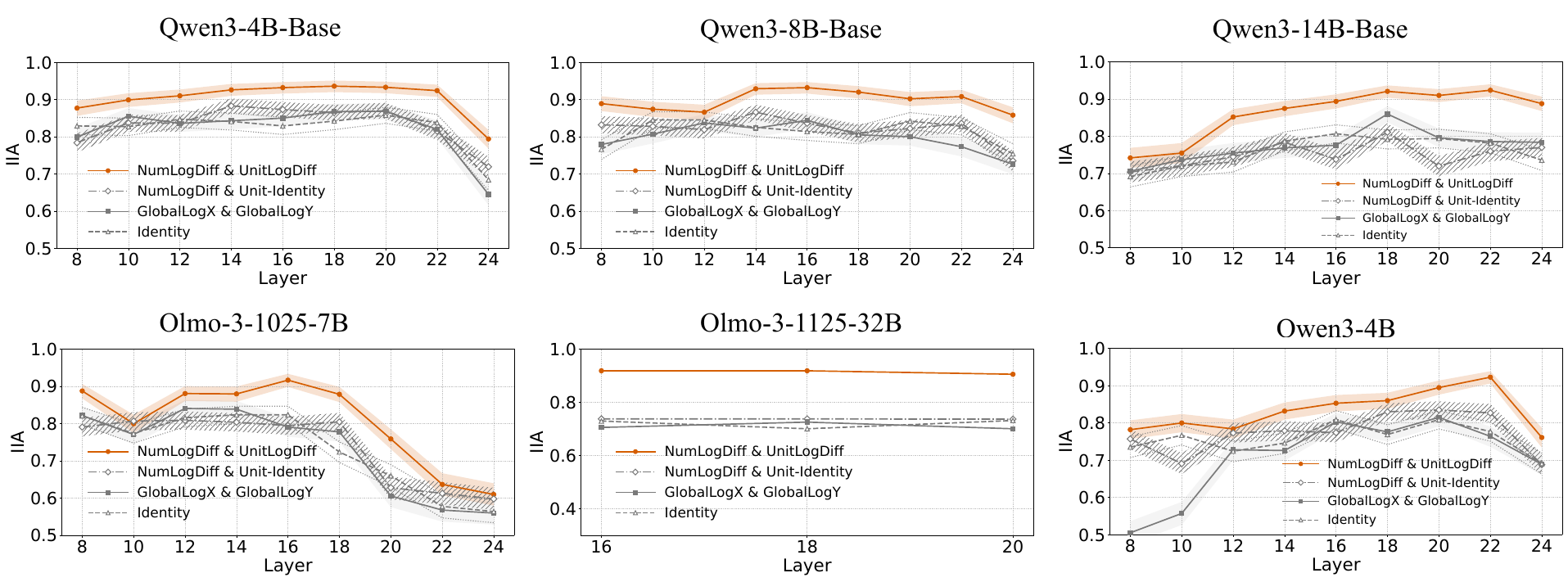}
\centering
\caption{
Layer-wise DAS intervention accuracy at the last token of u2 for Qwen3-4B-Base, Qwen3-8B-Base, Qwen3-14B-Base, Olmo-3-1025-7B, Olmo-3-1125-32B and an instruct-tuned LM (Qwen3-4B) on Metric length comparisons.
}
\label{fig:layerwise_iia_modelwise}
\end{figure*}

We first examine whether the main DAS result is robust across LMs.
\cref{fig:layerwise_iia_modelwise} shows layer-wise IIA at the last token of \(u_2\) for Qwen3-4B-Base, Qwen3-8B-Base, Qwen3-14B-Base, Olmo-3-1025-7B, Olmo-3-1125-32B, and an instruction-tuned LM (Qwen3-4B) on metric length comparisons.
Across LMs, the \(\mathrm{NumLogDiff}/\mathrm{UnitLogDiff}\) setting achieves higher IIA than the \(\mathrm{GlobalLogX}/\mathrm{GlobalLogY}\) baseline, identity baselines, and \(\mathrm{NumLogDiff}\) \& Unit-Identity baseline across most layers.
The absolute IIA differs across LMs, but the relative ordering is stable: interventions on numerical-difference and unit-scale-difference subspaces have the strongest effect on the LM's counterfactual output.

\begin{figure*}[t!]
\includegraphics[width=\linewidth]{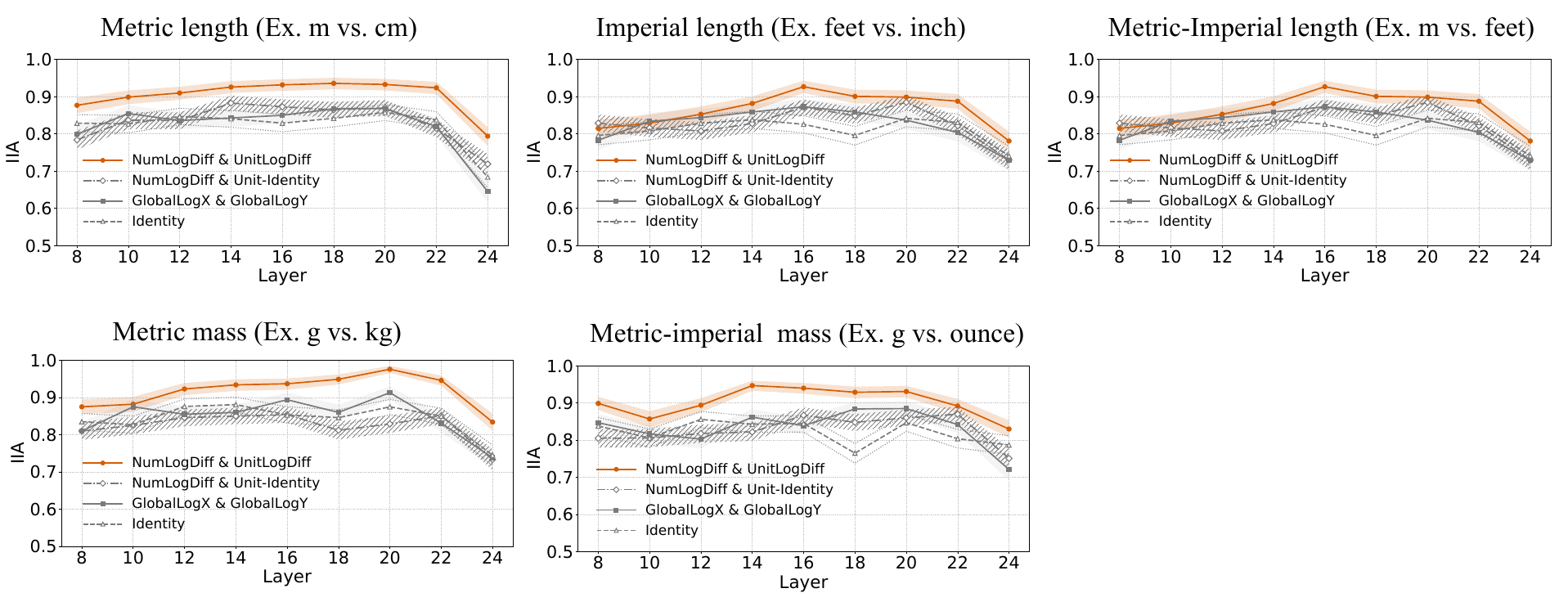}
\centering
\caption{
Layer-wise DAS intervention accuracy at the last token of \(u_2\) for Qwen3-4B-Base across all unit settings in \cref{tab:unit_settings}.
}
\label{fig:layerwise_iia_all_unitwise}
\end{figure*}

We next evaluate whether this pattern holds across unit settings.
\cref{fig:layerwise_iia_all_unitwise} reports the same layer-wise analysis for Qwen3-4B-Base across all unit settings in \cref{tab:unit_settings}.
The \(\mathrm{NumLogDiff}/\mathrm{UnitLogDiff}\) setting remains consistently above the two baselines across metric length, imperial length, metric-imperial length, metric mass, and metric-imperial mass settings.
This shows that the causal effect of numerical-difference and unit-scale-difference variables is not specific to one unit system or physical dimension.

\begin{figure*}[t!]
\includegraphics[width=\linewidth]{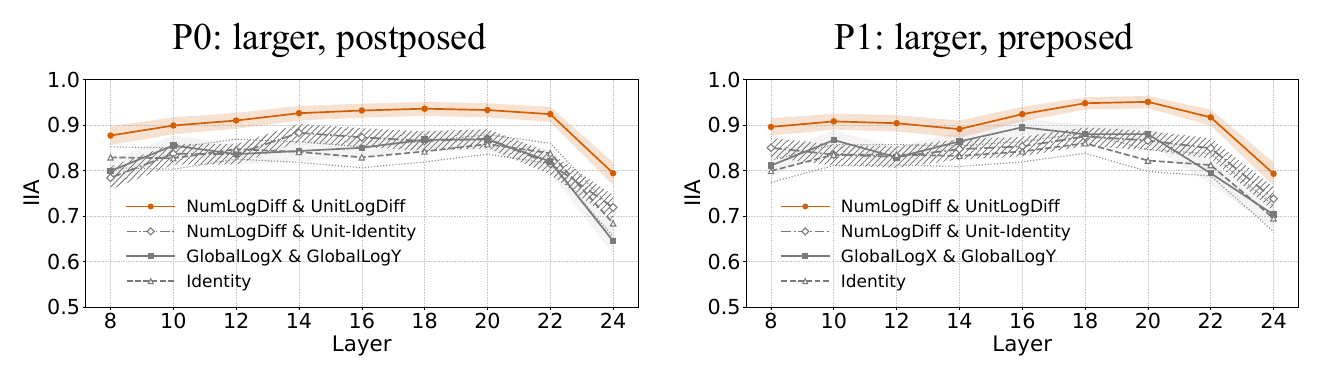}
\centering
\caption{
Layer-wise DAS intervention accuracy at the last token of \(u_2\) for Qwen3-4B-Base across Metric-mass settings under two prompt templates.
P0 is the larger-comparison postposed prompt, where the quantities appear after the comparison phrase, e.g., ``Which is larger, \(q_1\) or \(q_2\)?''
P1 is the larger-comparison preposed prompt, where the quantities appear before the comparison phrase, e.g., ``Between \(q_1\) and \(q_2\), which is larger?''
}
\label{fig:layerwise_iia_promptwise}
\end{figure*}

Furthermore, we test whether the result depends on prompt wording.
\cref{fig:layerwise_iia_promptwise} compares the larger-comparison postposed prompt P0 and the larger-comparison preposed prompt P1.
In both prompt templates, the \(\mathrm{NumLogDiff}/\mathrm{UnitLogDiff}\) setting obtains higher IIA than the \(\mathrm{GlobalLogX}/\mathrm{GlobalLogY}\) and identity baselines across layers.

Finally, we examine the effect of the random seed used during DAS training. 
\cref{tab:seed_results} reports the intervention accuracy at the last token of \(u_2\) for Qwen3-4B-Base on metric length comparisons across different DAS training random seeds. We evaluate three random seeds, including seed 42 used in \cref{fig:das_qwen3-4b_meter}. 
Although the intervention accuracy varies slightly across random seeds, the superiority of the \(\mathrm{NumLogDiff}/\mathrm{UnitLogDiff}\) setting over the baselines observed in \cref{fig:das_qwen3-4b_meter} is consistently maintained.

\subsection{Robustness to Intervention Dimensionality}
\label{sec:appendix_das_dimensionality}
\begin{figure*}[t!]
\includegraphics[width=\linewidth]{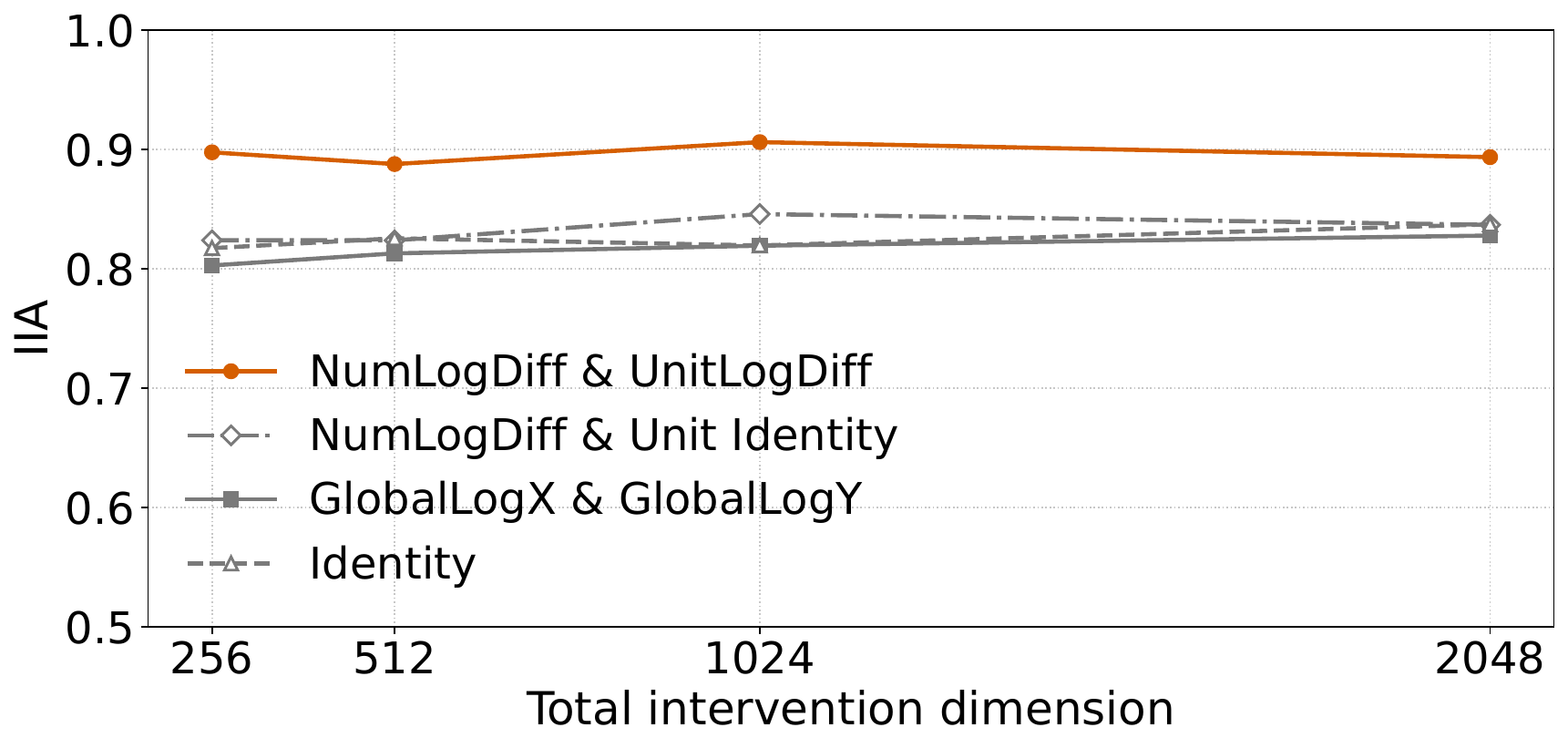}
\centering
\caption{
DAS intervention accuracy for Qwen3-4B-Base on the metric length setting as the total dimensionality of the non-residual intervention subspaces is varied.
The x-axis shows the total intervention dimensionality, which is evenly allocated across variables within each DAS setting.
The y-axis reports interchange intervention accuracy (IIA) at the last token of \(u_2\), averaged over the selected layers from 8 to 24 in steps of 2.
Across dimensionalities, the \(\mathrm{NumLogDiff}\)/\(\mathrm{UnitLogDiff}\) setting consistently achieves higher IIA than the \(\mathrm{GlobalLogX}\)/\(\mathrm{GlobalLogY}\), \(\mathrm{NumLogDiff}\) and Unit identity, and identity baselines.
}
\label{fig:dimwise_iia_meter}
\end{figure*}

The main DAS experiments fix the total dimensionality of the non-residual intervention subspaces to \(1024\), allocated evenly across variables.
To test whether the result depends on this dimensionality choice, we vary the total intervention dimensionality over \(256\), \(512\), \(1024\), and \(2048\) in the metric length setting.
For each DAS setting, the total dimensionality is evenly split among the intervened high-level variables.
Thus, the two-variable settings, \(\mathrm{NumLogDiff}/\mathrm{UnitLogDiff}\) and \(\mathrm{GlobalLogX}/\mathrm{GlobalLogY}\), use half of the total dimensionality for each variable, while the four-variable identity baseline uses one quarter of the total dimensionality for each primitive component.
\cref{fig:dimwise_iia_meter} shows IIA as a function of the total intervention dimensionality.
Across all dimensionalities, the \(\mathrm{NumLogDiff}/\mathrm{UnitLogDiff}\) setting consistently achieves higher IIA than the \(\mathrm{GlobalLogX}/\mathrm{GlobalLogY}\) and identity baselines.
Moreover, the relative ordering of the three settings remains stable as the dimensionality changes.
Furthermore, the NumLogDiff/UnitLogDiff setting with a total dimensionality of 1024 achieves higher IIA than the identity baseline with a total dimensionality of 2048, even though both settings allocate 512 dimensions per variable.
This indicates that the main causal result is not an artifact of a particular subspace size: numerical-difference and unit-scale-difference variables more directly steer the LM's comparison output across a range of intervention dimensionalities.

\subsection{Robustness to Counterfactual Examples}
\label{sec:das_matched_counterfactuals}

Different DAS settings define different interchange interventions, which may result in differences in the distributions and difficulty of counterfactual examples.
To examine whether such differences account for the observed differences in IIA, we conduct an additional experiment in which the base–target pairs are exactly matched across DAS settings.
We conduct this experiment on Qwen3-4B-Base in the metric length setting.

The IIA averaged over layers 18, 20, and 22 is 0.891 for NumLogDiff/UnitLogDiff, 0.805 for NumLogDiff and Unit identity, 0.808 for GlobalLogX/GlobalLogY, and 0.777 for the identity baseline.
The NumLogDiff/UnitLogDiff setting consistently achieves higher mean IIA than the alternative settings even when the base–target pairs are matched.
These results suggest that differences in counterfactual examples do not account for the superiority of the NumLogDiff/UnitLogDiff setting.

\subsection{Control Experiments for Learned Alignments}
\label{sec:das_alignment_controls}

To examine whether the observed IIA depends on the learned alignment between high-level variables and intervention subspaces, we conduct two control experiments for the NumLogDiff/UnitLogDiff setting with 512 dimensions per variable.
In the permuted-feature control, we swap the source inputs assigned to the two subspaces while keeping the learned rotation matrix fixed, thereby disrupting the correspondence between variables and subspaces.
In the random-rotation control, we use untrained random orthogonal rotation matrices and average the results over five random seeds.

The 95\% confidence interval bounds, averaged over layers 8 to 24 in steps of 2, are $[0.479, 0.541]$ and $[0.485, 0.547]$ for the permuted-feature and random-rotation controls, respectively.
Both controls yield near-chance IIA, suggesting that the high IIA observed in the main experiments depends on the learned alignment.

\begin{table*}[t]
\centering
\small
\begin{tabular}{llll}
\toprule
\textbf{Target} & \textbf{Dimension} & \textbf{Design} & \textbf{Prompt template} \\
\midrule
Larger & Length & Postposed &
\texttt{Which is larger, \(q_1\) or \(q_2\)? The answer is} \\
Larger & Length & Preposed &
\texttt{Between \(q_1\) and \(q_2\), which is larger? The answer is} \\
Larger & Mass & Postposed &
\texttt{Which is heavier, \(q_1\) or \(q_2\)? The answer is} \\
Larger & Mass & Preposed &
\texttt{Between \(q_1\) and \(q_2\), which is heavier? The answer is} \\
\midrule
Smaller & Length & Postposed &
\texttt{Which is smaller, \(q_1\) or \(q_2\)? The answer is} \\
Smaller & Length & Preposed &
\texttt{Between \(q_1\) and \(q_2\), which is smaller? The answer is} \\
Smaller & Mass & Postposed &
\texttt{Which is lighter, \(q_1\) or \(q_2\)? The answer is} \\
Smaller & Mass & Preposed &
\texttt{Between \(q_1\) and \(q_2\), which is lighter? The answer is} \\
\bottomrule
\end{tabular}
\caption{
Prompt templates used in the behavioral analysis.
In the postposed design, the two quantities are placed after the comparison phrase. 
In the preposed design, they are placed before the comparison phrase.
}
\label{tab:prompt_templates}
\end{table*}

\begin{table*}[t]
\centering
\small
\begin{tabular}{lll}
\toprule
\textbf{Unit} & \textbf{Short notation} & \textbf{Long notation} \\
\midrule
Millimeter & \texttt{mm} & \texttt{millimeter} \\
Centimeter & \texttt{cm} & \texttt{centimeter} \\
Meter & \texttt{m} & \texttt{meter} \\
Kilometer & \texttt{km} & \texttt{kilometer} \\
\midrule
Inch & \texttt{in} & \texttt{inch} \\
Foot & \texttt{ft} & \texttt{foot} \\
Yard & \texttt{yd} & \texttt{yard} \\
Mile & \texttt{mi} & \texttt{mile} \\
\midrule
Milligram & \texttt{mg} & \texttt{milligram} \\
Gram & \texttt{g} & \texttt{gram} \\
Kilogram & \texttt{kg} & \texttt{kilogram} \\
Tonne & \texttt{t} & \texttt{tonne} \\
\midrule
Ounce & \texttt{oz} & \texttt{ounce} \\
Pound & \texttt{lb} & \texttt{pound} \\
\bottomrule
\end{tabular}
\caption{
Short and long unit notations used in the experiments.
}
\label{tab:unit_notations}
\end{table*}

\begin{table*}[t]
\centering
\small
\setlength{\tabcolsep}{4pt}
\begin{tabular}{l|cccccccc}
\toprule
\textbf{Answer format} &
\multicolumn{8}{c}{\textbf{Quantity Margin (QM)}} \\
& $(-2,-1]$ & $(-1,-0.5]$ &
$(-0.5,-0.2]$ & $(-0.2,0]$ &
$(0,0.2]$ & $(0.2,0.5]$ &
$(0.5,1]$ & $(1,2]$ \\
\midrule
Full strings
& 1.00 & 0.93 & 0.88 & 0.79
& 0.75 & 0.87 & 0.92 & 1.00 \\
A/B choice
& 1.00 & 0.97 & 0.91 & 0.82
& 0.79 & 0.93 & 0.99 & 1.00 \\
\bottomrule
\end{tabular}
\caption{
Comparison accuracy across Quantity Margin intervals
for two answer formats: full answer strings and A/B choices.
Both answer formats exhibit a decrease in accuracy near
the comparison boundary.
}
\label{tab:behavioral_answer_format}
\end{table*}

\begin{table*}[t]
\centering
\small
\begin{tabular}{p{0.2\linewidth}p{0.6\linewidth}}
\toprule
\textbf{Example} & \textbf{Representative output} \\
\midrule
\textbf{Difficult example}\\
4.597 cm vs.\ 46.431 mm
&
``1 cm is equal to 10 mm. (...) 4.597 cm becomes 45.97 mm. (...) Since 45.97 is less than 46.431, the mm value is larger. (...) After converting both values to the same unit and comparing, 46.431 mm is larger than 4.597 cm. Answer: 46.431 mm''
\\
\midrule
\textbf{Easy example}\\
53.358 cm vs.\ 0.047 mm
&
``Since 1 cm = 10 mm, 53.358 cm $\times$ 10 = 533.58 mm. (...) 0.047 mm is a tiny fraction of a millimeter. So 53.358 cm is way larger. (...) Answer: 53.358 cm''
\\
\bottomrule
\end{tabular}
\caption{Below are representative Qwen3-4B-Thinking-2507 outputs; omitted portions are marked by ``(...)''.}
\label{tab:thinking_examples}
\end{table*}

\begin{table*}[t]
\centering
\small
\begin{tabular}{lcccc}
\toprule
 & Layer 8 & Layer 12 & Layer 16 & Layer 20 \\
\midrule
Seed 0  & 0.876 & 0.893 & 0.945 & 0.955 \\
Seed 1  & 0.883 & 0.916 & 0.928 & 0.917 \\
Seed 42 & 0.879 & 0.913 & 0.939 & 0.935 \\
\bottomrule
\end{tabular}
\caption{Layer-wise DAS intervention accuracy at the
last token of u2 for Qwen3-4B-Base on metric length
comparisons across DAS training random seeds.}
\label{tab:seed_results}
\end{table*}

\clearpage

\clearpage
\begin{table*}[t]
\centering
\small
\begin{tabular}{lll}
\toprule
\textbf{Feature group} & \textbf{Feature} & \textbf{Definition / Interpretation} \\
\midrule

\multicolumn{3}{l}{\textbf{Continuous quantity features}} \\
\midrule
Left number log
& \(F_{1\mathrm{NumLog}}\)
& \(\log r_1\). \\

Right number log
& \(F_{2\mathrm{NumLog}}\)
& \(\log r_2\). \\

Left unit-scale log
& \(F_{1\mathrm{UnitLog}}\)
& \(\log s_D(u_1)\). \\

Right unit-scale log
& \(F_{2\mathrm{UnitLog}}\)
& \(\log s_D(u_2)\). \\

Number-log difference
& \(F_{\mathrm{NumLogDiff}}\)
& \(\log r_1 - \log r_2\). \\

Unit-log difference
& \(F_{\mathrm{UnitLogDiff}}\)
& \(\log s_D(u_1) - \log s_D(u_2)\). \\

Left global log
& \(F_{\mathrm{GlobalLogX}}\)
& \(\log r_1 + \log s_D(u_1)\). \\

Right global log
& \(F_{\mathrm{GlobalLogY}}\)
& \(\log r_2 + \log s_D(u_2)\). \\

Quantity Margin
& \(F_{\mathrm{QuantityMargin}}\)
& \(QM(q_1,q_2)\). \\

Left value in right unit
& \(F_{\mathrm{XAsYUnitLog}}\)
& \(\log r_1 + \log s_D(u_1) - \log s_D(u_2)\). \\

Right value in left unit
& \(F_{\mathrm{YAsXUnitLog}}\)
& \(\log r_2 + \log s_D(u_2) - \log s_D(u_1)\). \\

\addlinespace
\midrule
\multicolumn{3}{l}{\textbf{Signed comparison heuristics}} \\
\midrule
Number-log comparison
& \(H_{\mathrm{numlogdiff}}^{\pm}\)
& \(\mathrm{sign}(\log r_1 - \log r_2)\). \\

Unit-scale comparison
& \(H_{\mathrm{unitdiff}}^{\pm}\)
& \(\mathrm{sign}(\log s_D(u_1) - \log s_D(u_2))\). \\

\addlinespace
\midrule
\multicolumn{3}{l}{\textbf{Number threshold heuristics}} \\
\midrule
Left-number threshold
& \(H_{r_1 > 10^n}\)
& \(\mathbf{1}[\log r_1 > n]\), where \(n \in \{-3,-2,-1,0,1,2,3\}\). \\

Right-number threshold
& \(H_{r_2 > 10^n}\)
& \(\mathbf{1}[\log r_2 > n]\), where \(n \in \{-3,-2,-1,0,1,2,3\}\). \\

\addlinespace
\midrule
\multicolumn{3}{l}{\textbf{Unit-scale threshold heuristics}} \\
\midrule
Left-unit threshold
& \(H_{s_D(u_1) > 10^n}\)
& \(\mathbf{1}[\log s_D(u_1) > n]\). \\

Right-unit threshold
& \(H_{s_D(u_2) > 10^n}\)
& \(\mathbf{1}[\log s_D(u_2) > n]\). \\

\addlinespace
\midrule
\multicolumn{3}{l}{\textbf{Global quantity threshold heuristics}} \\
\midrule
Left-global threshold
& \(H_{\mathrm{GlobalLogX} > n}\)
& \(\mathbf{1}[\log r_1 + \log s_D(u_1) > n]\). \\

Right-global threshold
& \(H_{\mathrm{GlobalLogY} > n}\)
& \(\mathbf{1}[\log r_2 + \log s_D(u_2) > n]\). \\

\addlinespace
\midrule
\multicolumn{3}{l}{\textbf{Unit-normalized threshold heuristics}} \\
\midrule
Left as right-unit threshold
& \(H_{\mathrm{XAsYUnitLog} > n}\)
& \(\mathbf{1}[\log r_1 + \log s_D(u_1) - \log s_D(u_2) > n]\). \\

Right as left-unit threshold
& \(H_{\mathrm{YAsXUnitLog} > n}\)
& \(\mathbf{1}[\log r_2 + \log s_D(u_2) - \log s_D(u_1) > n]\). \\

\addlinespace
\midrule
\multicolumn{3}{l}{\textbf{Difference threshold heuristics}} \\
\midrule
Number-log-difference threshold
& \(H_{\mathrm{NumLogDiff} > n}\)
& \(\mathbf{1}[\log r_1 - \log r_2 > n]\). \\

Unit-log-difference threshold
& \(H_{\mathrm{UnitLogDiff} > n}\)
& \(\mathbf{1}[\log s_D(u_1) - \log s_D(u_2) > n]\). \\

\addlinespace
\midrule
\multicolumn{3}{l}{\textbf{Unit identity heuristics}} \\
\midrule
Left-unit identity
& \(H_{u_1=u}\)
& \(\mathbf{1}[u_1=u]\). \\

Right-unit identity
& \(H_{u_2=u}\)
& \(\mathbf{1}[u_2=u]\). \\

\addlinespace
\midrule
\multicolumn{3}{l}{\textbf{Unit-pair heuristics}} \\
\midrule
Unit-pair identity
& \(H_{(u_1,u_2)=(u,v)}\)
& \(\mathbf{1}[(u_1,u_2)=(u,v)]\). \\

\bottomrule
\end{tabular}
\caption{
Continuous and heuristic features used in the surrogate models.
Here, \(q_1=r_1u_1\) and \(q_2=r_2u_2\), and \(s_D(u)\) denotes the scale of unit \(u\) relative to the base unit for physical dimension \(D\).
The continuous features preserve log-scale magnitudes and differences, while the heuristic features encode signed comparisons, threshold indicators, and unit-identity information.
\(F_{\mathrm{QuantityMargin}}\) corresponds to the correct comparison rule and is treated as a control feature rather than a surface heuristic.
Superscript \(\pm\) denotes signed comparison heuristics.
}
\label{tab:heuristic_features}
\end{table*}

\begin{table*}[t]
\centering
\small
\begin{tabular}{llp{0.4\linewidth}}
\toprule
\textbf{Model} & \textbf{Feature set} & \textbf{Included features} \\
\midrule

M0
& Intercept only
& Constant baseline with no input feature. \\

\midrule
\multicolumn{3}{l}{\textit{Correct-rule control}} \\

M1
& Quantity Margin only
& \(QM(q_1,q_2)\). \\

\midrule
\multicolumn{3}{l}{\textit{Continuous and algorithmic representations}} \\

M2
& NumLogDiff \& UnitLogDiff
& Continuous difference features:
\(\mathrm{NumLogDiff}\) and \(\mathrm{UnitLogDiff}\). \\

M3
& GlobalLogX \& GlobalLogY
& Continuous global-magnitude features:
\(\mathrm{GlobalLogX}\) and \(\mathrm{GlobalLogY}\). \\

M4
& Left-unit normalized
& Left-unit comparison features:
\(\log r_1\) and \(\mathrm{YAsXUnitLog}\). \\

M5
& Right-unit normalized
& Right-unit comparison features:
\(\mathrm{XAsYUnitLog}\) and \(\log r_2\). \\

M6
& \(r_1\), \(r_2\), \(u_1\), \(u_2\) logs
& Primitive log-space variables:
\(\log r_1\),
\(\log r_2\),
\(\log s_D(u_1)\), and
\(\log s_D(u_2)\). \\

M7
& All continuous core
& All continuous variables except direct \(QM(q_1,q_2)\). \\

\midrule
\multicolumn{3}{l}{\textit{Simple comparison heuristics}} \\

M8
& Signed NumLogDiff only
& Signed heuristic:
\(H_{\mathrm{NumLogDiff}}^{\pm}
=\mathrm{sign}(\mathrm{NumLogDiff})\). \\

M9
& Signed UnitLogDiff only
& Signed heuristic:
\(H_{\mathrm{UnitLogDiff}}^{\pm}
=\mathrm{sign}(\mathrm{UnitLogDiff})\). \\

M10
& Signed NumLogDiff \& UnitLogDiff
& Signed heuristics:
\(H_{\mathrm{NumLogDiff}}^{\pm}\) and
\(H_{\mathrm{UnitLogDiff}}^{\pm}\). \\

\midrule
\multicolumn{3}{l}{\textit{Identity heuristic families}} \\

M11
& Unit identity
& One-hot indicators for \(u_1\) and \(u_2\). \\

M12
& Unit-pair identity
& One-hot indicators for ordered unit pairs \((u_1,u_2)\). \\

\midrule
\multicolumn{3}{l}{\textit{Single-variable threshold heuristic families}} \\

M13
& \(r_1\) thresholds
& Threshold indicators over \(r_1\):
\(H_{r_1>10^n}\). \\

M14
& \(r_2\) thresholds
& Threshold indicators over \(r_2\):
\(H_{r_2>10^n}\). \\

M15
& \(u_1\)-scale thresholds
& Threshold indicators over the left unit scale:
\(H_{s_D(u_1)>10^n}\), equivalently
\(\mathbf{1}[\log s_D(u_1)>n]\). \\

M16
& \(u_2\)-scale thresholds
& Threshold indicators over the right unit scale:
\(H_{s_D(u_2)>10^n}\), equivalently
\(\mathbf{1}[\log s_D(u_2)>n]\). \\

M17
& GlobalLogX thresholds
& Threshold indicators over \(\mathrm{GlobalLogX}\). \\

M18
& GlobalLogY thresholds
& Threshold indicators over \(\mathrm{GlobalLogY}\). \\

M19
& X-as-Y-unit thresholds
& Threshold indicators over \(\mathrm{XAsYUnitLog}\). \\

M20
& Y-as-X-unit thresholds
& Threshold indicators over \(\mathrm{YAsXUnitLog}\). \\

M21
& NumLogDiff thresholds
& Threshold indicators over \(\mathrm{NumLogDiff}\). \\

M22
& UnitLogDiff thresholds
& Threshold indicators over \(\mathrm{UnitLogDiff}\). \\

\midrule
\multicolumn{3}{l}{\textit{Grouped threshold heuristic families}} \\

M23
& Number thresholds
& Both \(r_1\)- and \(r_2\)-threshold features. \\

M24
& Unit-scale thresholds
& Both \(s_D(u_1)\)- and \(s_D(u_2)\)-threshold features. \\

M25
& Primitive component thresholds
& Threshold features over \(r_1\), \(r_2\), \(s_D(u_1)\), and \(s_D(u_2)\). \\

M26
& GlobalLogX \& GlobalLogY thresholds
& Both global-magnitude threshold families. \\

M27
& XAsYUnitLog \& YAsXUnitLog thresholds
& Both unit-normalized threshold families. \\

M28
& NumLogDiff \& UnitLogDiff thresholds
& Threshold features over the two difference variables. \\

M29
& All threshold heuristics
& All threshold features from M13--M22. \\

\midrule
\multicolumn{3}{l}{\textit{Combined heuristic models}} \\

M30
& All heuristic features
& Signed comparison heuristics from M10, identity heuristics from M11--M12, and all threshold heuristics from M29. \\

\midrule
\multicolumn{3}{l}{\textit{unit-identity and unit-pair baselines} for M28} \\

M31
& NumLogDiff thresholds + unit-pair thresholds
& NumLogDiff threshold features together with the threshold features for both unit scales. \\

M32
& NumLogDiff thresholds + \(u_1\)-unit thresholds
& NumLogDiff threshold features together with the left-unit-scale threshold features. \\

M33
& NumLogDiff thresholds + \(u_2\)-unit thresholds
& NumLogDiff threshold features together with the right-unit-scale threshold features. \\

\bottomrule
\end{tabular}
\caption{
Feature sets for the surrogate models.
The response variable is the LM's mean log-probability margin between the left and right candidate answers.
Here, \(q_1=r_1u_1\), \(q_2=r_2u_2\), and \(s_D(u)\) denotes the scale of unit \(u\) relative to the base unit for physical dimension \(D\).
}
\label{tab:surrogate_feature_sets}
\end{table*}

\begin{table*}[t]
\centering
\small
\begin{tabular}{ll|rrr|rrr}
\toprule
\multirow{2}{*}{\textbf{Model}} & \multirow{2}{*}{\textbf{Feature set}} &
\multicolumn{3}{c|}{\textbf{All examples}} &
\multicolumn{3}{c}{\textbf{Boundary examples}} \\
&
& \(R^2_{\mathrm{S}}\) & \(\Delta R^2\) & \(R^2_{\mathrm{partial}}\)
& \(R^2_{\mathrm{S}}\) & \(\Delta R^2\) & \(R^2_{\mathrm{partial}}\) \\
\midrule
M30 & All heuristic features & 0.886 & 0.196 & 0.634 & 0.687 & 0.666 & 0.680 \\
M28 & NumLogDiff \& UnitLogDiff thresholds & 0.836 & 0.147 & 0.472 & 0.749 & 0.728 & 0.740 \\
M10 & Signed NumLogDiff \& UnitLogDiff & 0.817 & 0.128 & 0.439 & 0.763 & 0.741 & 0.695 \\
M31 & NumLogDiff thresholds + unit-pair thresholds & 0.815 & 0.125 & 0.403 & 0.693 & 0.672 & 0.683 \\
M32 & NumLogDiff thresholds + \(u_1\)-unit thresholds & 0.754 & 0.064 & 0.395 & 0.550 & 0.529 & 0.654 \\
M33 & NumLogDiff thresholds + \(u_2\)-unit thresholds & 0.726 & 0.036 & 0.391 & 0.444 & 0.423 & 0.645 \\
M21 & NumLogDiff thresholds & 0.643 & -0.047 & 0.386 & 0.345 & 0.324 & 0.641 \\
M29 & All threshold heuristics & 0.792 & 0.102 & 0.344 & 0.075 & 0.053 & 0.080 \\
M27 & XAsYUnitLog \& YAsXUnitLog thresholds & 0.626 & -0.063 & 0.314 & -0.359 & -0.381 & 0.047 \\
M20 & Y-as-X-unit thresholds & 0.555 & -0.135 & 0.197 & -0.353 & -0.375 & -0.005 \\
M19 & X-as-Y-unit thresholds & 0.569 & -0.121 & 0.146 & -0.270 & -0.291 & 0.001 \\
M12 & Unit-pair identity & 0.377 & -0.313 & 0.114 & -0.437 & -0.458 & 0.053 \\
M22 & UnitLogDiff thresholds & 0.377 & -0.313 & 0.114 & -0.471 & -0.492 & 0.049 \\
M25 & Primitive component thresholds & 0.714 & 0.025 & 0.091 & 0.077 & 0.055 & 0.072 \\
M26 & GlobalLogX \& GlobalLogY thresholds & 0.713 & 0.023 & 0.087 & 0.027 & 0.006 & 0.024 \\
M23 & Number thresholds & 0.552 & -0.138 & 0.081 & -0.236 & -0.258 & 0.047 \\
M24 & Unit-scale thresholds & 0.376 & -0.314 & 0.081 & -0.480 & -0.502 & 0.065 \\
M11 & Unit identity & 0.376 & -0.314 & 0.081 & -0.464 & -0.486 & 0.065 \\
M7 & All continuous core & 0.714 & 0.024 & 0.079 & 0.077 & 0.055 & 0.056 \\
M6 & \(r_1\), \(r_2\), \(u_1\), \(u_2\) logs & 0.714 & 0.025 & 0.078 & 0.076 & 0.055 & 0.056 \\
M8 & Signed NumLogDiff only & 0.553 & -0.137 & 0.072 & -0.195 & -0.216 & 0.042 \\
M9 & Signed UnitLogDiff only & 0.373 & -0.317 & 0.072 & -0.503 & -0.524 & 0.043 \\
M2 & NumLogDiff \& UnitLogDiff & 0.712 & 0.022 & 0.071 & 0.063 & 0.042 & 0.043 \\
M16 & \(u_2\)-scale thresholds & 0.196 & -0.494 & 0.044 & -0.524 & -0.546 & 0.049 \\
M17 & GlobalLogX thresholds & 0.488 & -0.201 & 0.038 & -0.279 & -0.301 & 0.019 \\
M4 & Left-unit normalized & 0.701 & 0.012 & 0.038 & 0.052 & 0.030 & 0.031 \\
M13 & \(r_1\) thresholds & 0.339 & -0.350 & 0.037 & -0.298 & -0.320 & 0.030 \\
M18 & GlobalLogY thresholds & 0.426 & -0.264 & 0.035 & -0.485 & -0.506 & -0.006 \\
M14 & \(r_2\) thresholds & 0.316 & -0.373 & 0.026 & -0.361 & -0.382 & 0.011 \\
M5 & Right-unit normalized & 0.695 & 0.005 & 0.016 & 0.028 & 0.006 & 0.006 \\
M15 & \(u_1\)-scale thresholds & 0.244 & -0.445 & 0.009 & -0.386 & -0.407 & 0.003 \\
M3 & GlobalLogX \& GlobalLogY & 0.692 & 0.002 & 0.007 & 0.035 & 0.013 & 0.013 \\
M0 & Intercept only & 0.000 & -0.690 & 0.000 & -0.004 & -0.026 & 0.000 \\
M1 & Quantity Margin & 0.690 & 0.000 & 0.000 & 0.022 & 0.000 & 0.000 \\
\bottomrule
\end{tabular}
\caption{Detailed out-of-sample \(R^2\) comparison for the metric length setting. The same trained surrogate models are evaluated on All examples and Boundary examples. \(R^2_{\mathrm{S}}\) is the test \(R^2\) obtained from the current surrogate feature set. \(\Delta R^2=R^2_{\mathrm{S}}-R^2_{\mathrm{QM}}\), and \(R^2_{\mathrm{partial}}=(R^2_{\mathrm{QM+S}}-R^2_{\mathrm{QM}})/(1-R^2_{\mathrm{QM}})\). Rows are sorted by \(R^2_{\mathrm{partial}}\) on All examples.}
\label{tab:meter_surrogate_detailed_results}
\end{table*}

\begin{table*}[t]
\centering
\small
\begin{tabular}{l|rr|rr}
\toprule
\multirow{2}{*}{\textbf{Feature set}} &
\multicolumn{2}{c|}{\textbf{MLP}} &
\multicolumn{2}{c}{\textbf{Sigmoid}} \\
& All & Boundary & All & Boundary \\
\midrule
NumLogDiff \& UnitLogDiff & 0.957 & 0.736 & 0.816 & 0.731 \\
GlobalLogX \& GlobalLogY & 0.873 & 0.359 & 0.712 & 0.038 \\
Quantity Margin & 0.865 & 0.356 & 0.827 & 0.091 \\
\bottomrule
\end{tabular}
\caption{
Nonlinear surrogate results for the metric length setting in Qwen3-4B-Base.
For each feature set, we evaluate one-hidden-layer MLPs and sigmoid-based surrogate models, with their configurations selected on the validation set.
We report the test $R^2$ on All examples and Boundary examples.
}
\label{tab:nonlinear_surrogate_results}
\end{table*}

\begin{table*}[t]
\centering
\small
\begin{tabular}{l|rr}
\toprule
\multirow{2}{*}{\textbf{Feature set}} &
\multicolumn{2}{c}{\textbf{$R^2$}} \\
& All & Boundary \\
\midrule
NumLogDiff \& UnitLogDiff thresholds
& 0.836 & 0.749 \\
FirstDigitDiff \& DigitCountDiff \& UnitLogDiff thresholds
& 0.720 & 0.160 \\
DigitCountDiff \& UnitLogDiff thresholds
& 0.668 & -0.112 \\
FirstDigitDiff \& UnitLogDiff thresholds
& 0.630 & 0.186 \\
\bottomrule
\end{tabular}
\caption{
Surrogate results for digit-based controls in the metric
length setting in Qwen3-4B-Base.
We report the test $R^2$ on All examples and Boundary examples.
The four feature sets contain 38, 49, 31, and 37 features,
respectively.
}
\label{tab:digit_control_surrogate_results}
\end{table*}

\begin{table*}[t]
\centering
\small
\begin{tabular}{l|rrr|rrr}
\toprule
\multirow{2}{*}{\textbf{Feature set}} &
\multicolumn{3}{c|}{\textbf{All examples}} &
\multicolumn{3}{c}{\textbf{Boundary examples}} \\
& \(R^2_{\mathrm{S}}\) & \(\Delta R^2\) & \(R^2_{\mathrm{partial}}\)
& \(R^2_{\mathrm{S}}\) & \(\Delta R^2\) & \(R^2_{\mathrm{partial}}\) \\
\midrule
All heuristic features & 0.886 & 0.196 & 0.634 & 0.687 & 0.666 & 0.680 \\
NumLogDiff \& UnitLogDiff thresholds & 0.836 & 0.147 & 0.472 & 0.749 & 0.728 & 0.740 \\
Signed NumLogDiff \& UnitLogDiff & 0.817 & 0.128 & 0.439 & 0.763 & 0.741 & 0.695 \\
NumLogDiff thresholds + unit-pair thresholds & 0.815 & 0.125 & 0.403 & 0.693 & 0.672 & 0.683 \\
Primitive component thresholds & 0.714 & 0.025 & 0.091 & 0.077 & 0.055 & 0.072 \\
NumLogDiff \& UnitLogDiff & 0.712 & 0.022 & 0.071 & 0.063 & 0.042 & 0.043 \\
GlobalLogX \& GlobalLogY & 0.692 & 0.002 & 0.007 & 0.035 & 0.013 & 0.013 \\
Quantity Margin & 0.690 & 0.000 & 0.000 & 0.022 & 0.000 & 0.000 \\
\bottomrule
\end{tabular}
\caption{
Representative surrogate results for the metric length setting in Qwen3-4B-Base. 
The trained surrogate models are evaluated on All examples and Boundary examples. 
\(R^2_{\mathrm{QM}}\), \(R^2_{\mathrm{S}}\), and \(R^2_{\mathrm{QM+S}}\) denote the test \(R^2\) of the Quantity Margin baseline, the current surrogate model, and the surrogate model obtained by adding the Quantity Margin to the current feature set, respectively.
\(\Delta R^2=R^2_{\mathrm{S}}-R^2_{\mathrm{QM}}\), and \(R^2_{\mathrm{partial}}=(R^2_{\mathrm{QM+S}}-R^2_{\mathrm{QM}})/(1-R^2_{\mathrm{QM}})\). 
Rows are sorted by \(R^2_{\mathrm{partial}}\) on All examples.
}
\label{tab:meter_test_quantity_margin_abs_lt_0_2_r2_representative}
\end{table*}

\begin{table*}[t]
\centering
\small
\begin{tabular}{l|rrr|rrr}
\toprule
\multirow{2}{*}{\textbf{Feature set}} &
\multicolumn{3}{c|}{\textbf{All examples}} &
\multicolumn{3}{c}{\textbf{Boundary examples}} \\
&
\(R^2_{\mathrm{S}}\) & \(\Delta R^2\) & \(R^2_{\mathrm{partial}}\) &
\(R^2_{\mathrm{S}}\) & \(\Delta R^2\) & \(R^2_{\mathrm{partial}}\) \\
\midrule
All heuristic features & 0.871 & 0.224 & 0.635 & 0.650 & 0.630 & 0.639 \\
NumLogDiff \& UnitLogDiff thresholds & 0.822 & 0.176 & 0.497 & 0.643 & 0.622 & 0.625 \\
Signed NumLogDiff \& UnitLogDiff & 0.810 & 0.163 & 0.473 & 0.710 & 0.689 & 0.636 \\
NumLogDiff thresholds + unit-pair thresholds & 0.790 & 0.143 & 0.411 & 0.525 & 0.505 & 0.500 \\
Primitive component thresholds & 0.662 & 0.016 & 0.075 & 0.001 & -0.020 & 0.005 \\
NumLogDiff \& UnitLogDiff & 0.659 & 0.013 & 0.036 & 0.026 & 0.005 & 0.006 \\
Quantity Margin & 0.647 & 0.000 & 0.000 & 0.021 & 0.000 & 0.000 \\
GlobalLogX \& GlobalLogY & 0.646 & -0.001 & -0.002 & 0.017 & -0.003 & -0.004 \\
\bottomrule
\end{tabular}
\caption{
Representative surrogate results for the imperial length setting in Qwen3-4B-Base. 
The trained surrogate models are evaluated on All examples and Boundary examples. 
\(R^2_{\mathrm{QM}}\), \(R^2_{\mathrm{S}}\), and \(R^2_{\mathrm{QM+S}}\) denote the test \(R^2\) of the Quantity Margin baseline, the current surrogate model, and the surrogate model obtained by adding the Quantity Margin to the current feature set, respectively.
\(\Delta R^2=R^2_{\mathrm{S}}-R^2_{\mathrm{QM}}\), and \(R^2_{\mathrm{partial}}=(R^2_{\mathrm{QM+S}}-R^2_{\mathrm{QM}})/(1-R^2_{\mathrm{QM}})\). 
Rows are sorted by \(R^2_{\mathrm{partial}}\) on All examples.
}
\label{tab:feet_test_quantity_margin_abs_lt_0_2_r2_representative}
\end{table*}

\begin{table*}[t]
\centering
\small
\begin{tabular}{l|rrr|rrr}
\toprule
\multirow{2}{*}{\textbf{Feature set}} &
\multicolumn{3}{c|}{\textbf{All examples}} &
\multicolumn{3}{c}{\textbf{Boundary examples}} \\
&
\(R^2_{\mathrm{S}}\) & \(\Delta R^2\) & \(R^2_{\mathrm{partial}}\) &
\(R^2_{\mathrm{S}}\) & \(\Delta R^2\) & \(R^2_{\mathrm{partial}}\) \\
\midrule
All heuristic features & 0.938 & 0.161 & 0.722 & 0.636 & 0.693 & 0.658 \\
NumLogDiff \& UnitLogDiff thresholds & 0.851 & 0.075 & 0.354 & 0.303 & 0.360 & 0.362 \\
NumLogDiff thresholds + unit-pair thresholds & 0.838 & 0.061 & 0.304 & 0.259 & 0.316 & 0.350 \\
Primitive component thresholds & 0.819 & 0.042 & 0.232 & 0.251 & 0.309 & 0.342 \\
Signed NumLogDiff \& UnitLogDiff & 0.703 & -0.073 & 0.141 & -0.060 & -0.003 & 0.110 \\
NumLogDiff \& UnitLogDiff & 0.779 & 0.002 & 0.010 & -0.074 & -0.017 & -0.016 \\
GlobalLogX \& GlobalLogY & 0.777 & 0.001 & 0.002 & -0.057 & 0.000 & 0.000 \\
Quantity Margin & 0.777 & 0.000 & 0.000 & -0.057 & 0.000 & 0.000 \\
\bottomrule
\end{tabular}
\caption{
Representative surrogate results for the metric-imperial length setting in Qwen3-4B-Base. 
The trained surrogate models are evaluated on All examples and Boundary examples. 
\(R^2_{\mathrm{QM}}\), \(R^2_{\mathrm{S}}\), and \(R^2_{\mathrm{QM+S}}\) denote the test \(R^2\) of the Quantity Margin baseline, the current surrogate model, and the surrogate model obtained by adding the Quantity Margin to the current feature set, respectively.
\(\Delta R^2=R^2_{\mathrm{S}}-R^2_{\mathrm{QM}}\), and \(R^2_{\mathrm{partial}}=(R^2_{\mathrm{QM+S}}-R^2_{\mathrm{QM}})/(1-R^2_{\mathrm{QM}})\). 
Rows are sorted by \(R^2_{\mathrm{partial}}\) on All examples.
}
\label{tab:meter_feet_test_quantity_margin_abs_lt_0_2_r2_representative}
\end{table*}

\begin{table*}[t]
\centering
\small
\begin{tabular}{l|rrr|rrr}
\toprule
\multirow{2}{*}{\textbf{Feature set}} &
\multicolumn{3}{c|}{\textbf{All examples}} &
\multicolumn{3}{c}{\textbf{Boundary examples}} \\
&
\(R^2_{\mathrm{S}}\) & \(\Delta R^2\) & \(R^2_{\mathrm{partial}}\) &
\(R^2_{\mathrm{S}}\) & \(\Delta R^2\) & \(R^2_{\mathrm{partial}}\) \\
\midrule
All heuristic features & 0.833 & 0.321 & 0.660 & 0.632 & 0.616 & 0.624 \\
NumLogDiff \& UnitLogDiff thresholds & 0.736 & 0.224 & 0.460 & 0.742 & 0.725 & 0.754 \\
Signed NumLogDiff \& UnitLogDiff & 0.724 & 0.212 & 0.438 & 0.723 & 0.707 & 0.704 \\
NumLogDiff thresholds + unit-pair thresholds & 0.687 & 0.175 & 0.358 & 0.696 & 0.680 & 0.720 \\
Primitive component thresholds & 0.566 & 0.053 & 0.120 & -0.010 & -0.026 & -0.012 \\
NumLogDiff \& UnitLogDiff & 0.553 & 0.041 & 0.084 & -0.005 & -0.021 & -0.021 \\
GlobalLogX \& GlobalLogY & 0.518 & 0.005 & 0.011 & 0.016 & 0.000 & 0.000 \\
Quantity Margin & 0.512 & 0.000 & 0.000 & 0.016 & 0.000 & 0.000 \\
\bottomrule
\end{tabular}
\caption{
Representative surrogate results for the metric mass setting in Qwen3-4B-Base. 
The trained surrogate models are evaluated on All examples and Boundary examples. 
\(R^2_{\mathrm{QM}}\), \(R^2_{\mathrm{S}}\), and \(R^2_{\mathrm{QM+S}}\) denote the test \(R^2\) of the Quantity Margin baseline, the current surrogate model, and the surrogate model obtained by adding the Quantity Margin to the current feature set, respectively.
\(\Delta R^2=R^2_{\mathrm{S}}-R^2_{\mathrm{QM}}\), and \(R^2_{\mathrm{partial}}=(R^2_{\mathrm{QM+S}}-R^2_{\mathrm{QM}})/(1-R^2_{\mathrm{QM}})\). 
Rows are sorted by \(R^2_{\mathrm{partial}}\) on All examples.
}
\label{tab:gram_test_quantity_margin_abs_lt_0_2_r2_representative}
\end{table*}

\begin{table*}[t]
\centering
\small
\begin{tabular}{l|rrr|rrr}
\toprule
\multirow{2}{*}{\textbf{Feature set}} &
\multicolumn{3}{c|}{\textbf{All examples}} &
\multicolumn{3}{c}{\textbf{Boundary examples}} \\
&
\(R^2_{\mathrm{S}}\) & \(\Delta R^2\) & \(R^2_{\mathrm{partial}}\) &
\(R^2_{\mathrm{S}}\) & \(\Delta R^2\) & \(R^2_{\mathrm{partial}}\) \\
\midrule
All heuristic features & 0.873 & 0.198 & 0.611 & 0.671 & 0.660 & 0.666 \\
NumLogDiff thresholds + unit-pair thresholds & 0.826 & 0.151 & 0.468 & 0.635 & 0.625 & 0.619 \\
NumLogDiff \& UnitLogDiff thresholds & 0.815 & 0.139 & 0.429 & 0.591 & 0.581 & 0.587 \\
Signed NumLogDiff \& UnitLogDiff & 0.785 & 0.109 & 0.400 & 0.635 & 0.625 & 0.540 \\
Primitive component thresholds & 0.714 & 0.038 & 0.142 & 0.091 & 0.080 & 0.094 \\
NumLogDiff \& UnitLogDiff & 0.691 & 0.016 & 0.049 & 0.032 & 0.021 & 0.022 \\
GlobalLogX \& GlobalLogY & 0.677 & 0.001 & 0.004 & 0.015 & 0.005 & 0.005 \\
Quantity Margin & 0.675 & 0.000 & 0.000 & 0.010 & 0.000 & 0.000 \\
\bottomrule
\end{tabular}
\caption{
Representative surrogate results for the metric-imperial mass setting in Qwen3-4B-Base. 
The trained surrogate models are evaluated on All examples and Boundary examples. 
\(R^2_{\mathrm{QM}}\), \(R^2_{\mathrm{S}}\), and \(R^2_{\mathrm{QM+S}}\) denote the test \(R^2\) of the Quantity Margin baseline, the current surrogate model, and the surrogate model obtained by adding the Quantity Margin to the current feature set, respectively.
\(\Delta R^2=R^2_{\mathrm{S}}-R^2_{\mathrm{QM}}\), and \(R^2_{\mathrm{partial}}=(R^2_{\mathrm{QM+S}}-R^2_{\mathrm{QM}})/(1-R^2_{\mathrm{QM}})\). 
Rows are sorted by \(R^2_{\mathrm{partial}}\) on All examples.
}
\label{tab:gram_pound_test_quantity_margin_abs_lt_0_2_r2_representative}
\end{table*}

\end{document}